%% file: iclr2027_conference.tex
\documentclass{article}

\usepackage{iclr2027_conference,times}
\input{math_commands.tex}

\usepackage{graphicx}
\usepackage{booktabs}
\usepackage{multirow}
\usepackage{array}
\usepackage{amsmath,amssymb}
\usepackage{xcolor}
\usepackage{microtype}
\usepackage{hyperref}
\hypersetup{
  hidelinks,
  pdftitle={SPECTRA: Source-Preserving Editing with Cross-Scale Transform and Representation Adaptation},
  pdfauthor={Md Sakib Hossain Shovon, Dahee Kwon, Jaesik Choi},
  pdfsubject={Training-free source-preserving image editing},
  pdfkeywords={Image editing, diffusion models, flow matching, source preservation}
}
\usepackage{url}
\usepackage{xspace}
\usepackage{subcaption}
\usepackage{algorithm}
\usepackage{algpseudocode}
\usepackage{placeins}
\usepackage{float}

\title{SPECTRA: Source-Preserving Editing with Cross-Scale Transform and Representation Adaptation}

\author{\textbf{Md Sakib Hossain Shovon}, \quad \textbf{Dahee Kwon}, \quad \textbf{Jaesik Choi}\\
\normalfont KAIST AI, Daejeon, Republic of Korea\\
\normalfont\texttt{\{sakib, daheekwon, jaesik.choi\}@kaist.ac.kr}}

\iclrfinalcopy
\usepackage{etoolbox}
\makeatletter
\patchcmd{\@maketitle}
  {Published as a conference paper at ICLR 2027}
  {Preprint}
  {}
  {\PackageError{spectra-preprint}{Unable to set preprint header}{}}
\patchcmd{\@maketitle}
  {\hsize\textwidth}
  {\hsize\textwidth\centering}
  {}
  {\PackageError{spectra-preprint}{Unable to center title block}{}}
\patchcmd{\@maketitle}
  {\LARGE\sc \@title\par}
  {\LARGE\bfseries \@title\par}
  {}
  {\PackageError{spectra-preprint}{Unable to set title font}{}}
\patchcmd{\@maketitle}
  {\begin{tabular}[t]{l}\bf\rule{\z@}{24pt}\@author\end{tabular}}
  {\begin{tabular}[t]{c}\normalfont\rule{\z@}{24pt}\@author\end{tabular}}
  {}
  {\PackageError{spectra-preprint}{Unable to center authors}{}}
\patchcmd{\@maketitle}
  {\vskip 0.3in minus 0.1in}
  {\vskip\dimexpr0.3in-11pt\relax minus 0.1in}
  {}
  {\PackageError{spectra-preprint}{Unable to set title spacing}{}}
\makeatother

\newcommand{\method}{\textsc{Spectra}\xspace}
\newcommand{\ssmo}{\textsc{SSMO}\xspace}
\newcommand{\dcrci}{\textsc{DCRCI}\xspace}
\newcommand{\selfattnconv}{Self Attn-Conv\xspace}

\newcommand{\CLIPT}{\textsc{Clip-T}\xspace}

\begin{document}
\maketitle

\begin{abstract}
Text-guided image editing must follow the target instruction while retaining source information that should remain unchanged. Existing training-free methods use feature injection, attention control, or predefined frequency regions, but typically apply coarse or fixed preservation policies. We study source trajectories and find that preservation evidence differs across transform scales, is non-uniform across orientations within a scale, and changes over sampling time. \method therefore defines preservation over scale, direction, space, and time. At each native step, source-only multiscale transform energies determine threshold-free scale--direction weights, while target-token attention supplies spatial evidence through Semantic Structure Mass-Odds (\ssmo). The fused clean prediction is returned to the backbone's native DDIM or FlowMatch scheduler, enabling one interface across Stable Diffusion and transformer flow backbones; appearance-preservation tasks optionally use decoder-cycle residual chroma injection (\dcrci). Across two appearance-changing and two appearance-preserving benchmarks, \method shifts the preservation--editability operating point toward stronger source retention. Controlled analysis identifies a strong scale-depth effect and a smaller, conditional directional refinement: directional benefit tends to increase with source anisotropy, and correct band assignment improves DINO and Edge relative to shuffling the same weights.
\end{abstract}

\section{Introduction}
\label{sec:intro}

Text-guided image editing must change a source according to a target prompt without unnecessarily altering geometry, identity, background, or appearance. Training-free methods address this preservation--editability tension through inversion/attention control \citep{hertz2023prompt,mokady2023null}, feature injection \citep{tumanyan2023pnp}, spatial guidance \citep{mo2024freecontrol}, or feed-forward structure/appearance control \citep{lin2024ctrlx}. Frequency-aware methods instead manipulate spectral content: FBSDiff substitutes frequency bands \citep{gao2024fbsdiff}, FreeDiff progressively truncates frequencies \citep{wu2024freediff}, FSI-Edit injects high-frequency reconstruction features \citep{yang2025fsiedit}, and FGD separates low-frequency style from high-frequency structure guidance \citep{gao2025fgd}. These works establish spectral control as useful, but leave a finer question: \emph{how should preservation be allocated when source information is unevenly distributed inside the representation?}

Our starting observation is that source-preservation evidence is not uniform at a fixed frequency magnitude or even at a fixed transform scale. In a multiscale directional representation, different scales carry complementary source information, while the energy distribution over orientations within a scale can be strongly anisotropic. Moreover, this directional profile evolves along the native denoising or flow trajectory. The resulting preservation problem is therefore structured along four axes: \textbf{scale}, \textbf{direction}, \textbf{space}, and \textbf{time}. A fixed band selection cannot express this structure, and a scale-only policy ignores variation among directions that share the same scale.

\begin{figure}[t]
    \centering
    \includegraphics[width=\linewidth]{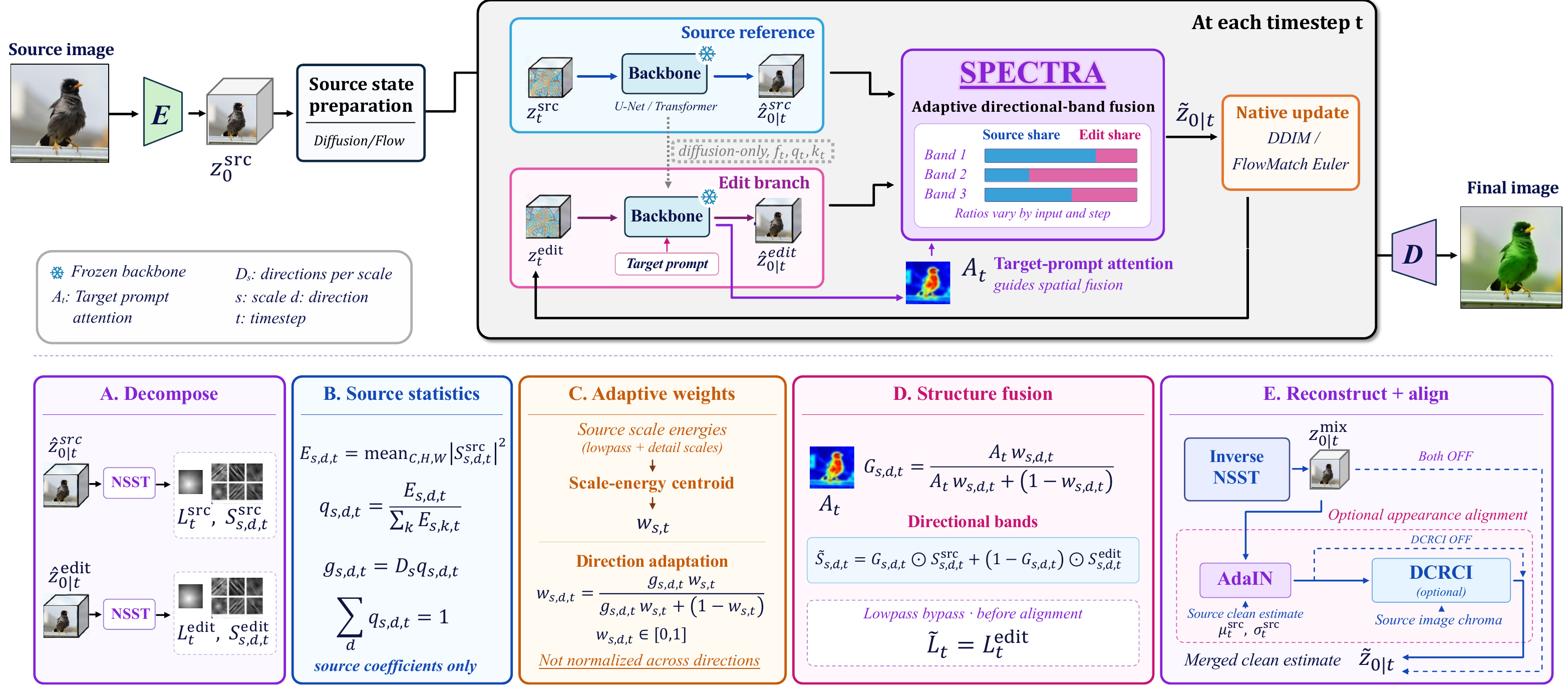}
    \vspace{-3pt}
    \caption{Overview of \method. Source-reference and target-conditioned clean predictions are decomposed at each native step. Source energies determine threshold-free scale--direction weights (\emph{what/when}); target-token attention supplies spatial evidence (\emph{where}) through \ssmo. The fused clean prediction is reconstructed, optionally aligned by AdaIN/\dcrci, and returned to the native DDIM or FlowMatch scheduler. \selfattnconv is U-Net-specific; clean-prediction fusion is shared across backbones.}
    \label{fig:spectra_main}
    \vspace{-10pt}
\end{figure}

Figure~\ref{fig:spectra_main} summarizes the per-step policy. A matched multiscale directional transform decomposes source-reference and target-conditioned clean predictions. Source coefficients alone determine scale and within-scale directional importance, without learned strengths, top-$k$ selection, frequency cutoffs, or direction-specific time windows. Native target attention instead answers \emph{where} this source prior should act through a continuous \ssmo gate. After reconstruction, the fused clean prediction is returned to the native scheduler. This interface is shared by DDIM-based Stable Diffusion \citep{rombach2022ldm,song2021ddim} and FlowMatch-based transformers \citep{lipman2023flowmatching,esser2024sd3,blackforestlabs2024flux}.

The results reveal a hierarchy. Increasing transform depth from one to three scales consistently improves source-preservation metrics on Style-9, supporting cross-scale complementarity. Direction-aware allocation is a smaller conditional refinement: its paired benefit tends to increase with source anisotropy, and a shuffled-direction control shows that correct band assignment gives the best DINO and Edge preservation. Thus the evidence links a measurable representation property and its assignment to when directional refinement is useful, without treating the anisotropy association as causal.

Our contributions are:
\begin{itemize}
    \item \textbf{Cross-scale directional heterogeneity.} We characterize source-preservation information as heterogeneous across transform scales, orientations within each scale, and sampling time. Scale-depth and trajectory analyses establish cross-scale complementarity and unequal, time-varying directional importance.
    \item \textbf{Threshold-free adaptive scale--direction preservation.} Source transform energies directly determine scale- and direction-specific weights without hand-tuned thresholds or subband selection. The rule reduces exactly to Scale-only under uniform directional energy, while a shuffled-direction control supports the importance of correct band assignment.
    \item \textbf{Semantic spatial gating and diffusion-internal color correction.} \ssmo couples source transform confidence with target-aware native attention to instantiate \emph{where} preservation acts. For appearance-preservation tasks, \dcrci provides targeted chroma correction; an isolated AdaIN-versus-AdaIN+\dcrci ablation improves CSD and $\Delta E_{00}$ while exposing its metric trade-off. The same clean-prediction interface applies across diffusion and flow backbones.
\end{itemize}

\section{Related Work}
\label{sec:related}

\paragraph{Training-free source-guided editing.}
Prompt-to-Prompt and Null-text inversion control attention or inversion states \citep{hertz2023prompt,mokady2023null}; PnP injects source spatial features/self-attention \citep{tumanyan2023pnp}; PnP Inversion introduces PIE-Bench \citep{ju2024pnpinversion}; FreeControl and Ctrl-X provide training-free spatial or structure/appearance control \citep{mo2024freecontrol,lin2024ctrlx}; LEDITS++ and DDPM-Edit offer localized editing and edit-friendly inversion \citep{brack2024ledits,huberman2024ddpmedit}. These methods preserve source information through inversion, feature reuse, attention, or spatial constraints. \method instead studies how preservation should be allocated within a multiscale directional representation as sampling evolves.

\paragraph{Frequency-aware and directional control.}
Frequency-conditioned editing includes learned DCT branches \citep{gao2024fcdiff}, band substitution in FBSDiff/FBSDiff++ \citep{gao2024fbsdiff,gao2026fbsdiffpp}, progressive truncation in FreeDiff \citep{wu2024freediff}, high-frequency suppression/injection in FlexiEdit and FSI-Edit \citep{koo2024flexiedit,yang2025fsiedit}, frequency-interactive attention in FIA-Edit \citep{yang2026fiaedit}, and low-/high-frequency style/structure guidance in FGD \citep{gao2025fgd}. Our claim is not that these representations lack orientation information; their control policies are primarily organized by predefined bands, radial regions, or fixed frequency roles. \method instead treats preservation as state-dependent allocation: the source determines scale/direction weights at each step, and native target attention determines where they act. Thus the contribution is not a new transform, but an adaptive preservation policy. NSST provides the default anisotropic multiscale representation \citep{easley2008shearlet}; Appendix~\ref{app:transforms} tests DTCWT, DWT, and NSCT.

W-Edit \citep{sun2026wedit} uses a cumulative-energy threshold to select wavelet subbands for source replacement in intermediate attention features. \method computes continuous source-derived scale--direction weights, uses target attention to modulate their spatial contribution, and fuses clean predictions before the native scheduler update. A matched comparison remains open because we could not locate a public author implementation.

\section{SPECTRA}
\label{sec:method}

\subsection{Problem Setup and Clean-Prediction Interface}
\label{sec:interface}

Let $x^{\mathrm{src}}$ be a source image and $p^{\mathrm{tgt}}$ a target editing prompt. A frozen VAE maps the source to a clean latent $z^{\mathrm{src}}_0$. At native sampling step $t$, the backbone produces a source-reference state $z_t^{\mathrm{src}}$ and an evolving target edit state $z_t^{\mathrm{edit}}$. The source branch is conditioned with empty text; the target prompt conditions only the edit branch. The scheduler-specific model outputs are converted to clean latent predictions
\begin{equation}
\hat z^{\mathrm{src}}_{0,t}, \qquad \hat z^{\mathrm{edit}}_{0,t}.
\label{eq:clean_preds}
\end{equation}
The spectral controller intervenes only on these clean predictions. It does not replace the denoiser, transformer, or scheduler.

For SD1.5/SD2.1, an empty-text DDIM inversion provides a source trajectory matched to the 50-step generation schedule. For SD3 and FLUX.1-dev, the source state is reconstructed at the selected native image-to-image timestep using the pipeline's scheduler-native noising operation and the same seed-fixed noise realization as the edit initialization. This separation preserves each backbone's native state construction while exposing the common clean-prediction interface used below.

Figure~\ref{fig:spectra_main} gives the complete per-step data flow. The key abstraction is the clean-prediction interface: backbone-specific state construction remains native, while the preservation policy acts on spatial clean latents before the scheduler update.

\subsection{Source-Conditioned Scale and Direction Allocation}
\label{sec:scale_direction}

\paragraph{Matched multiscale directional analysis.}
Let $\mathcal{T}$ denote a transform with terminal low-pass coefficient $L$ and directional detail coefficients $S_{s,d}$ at scale $s$ and direction $d$. We apply the same analysis to the two clean predictions:
\begin{align}
(L_t^{\mathrm{src}},\{S^{\mathrm{src}}_{s,d,t}\}) &= \mathcal{T}(\hat z^{\mathrm{src}}_{0,t}),\\
(L_t^{\mathrm{edit}},\{S^{\mathrm{edit}}_{s,d,t}\}) &= \mathcal{T}(\hat z^{\mathrm{edit}}_{0,t}).
\label{eq:transform}
\end{align}
Our default is a three-scale NSST with the \texttt{dmaxflat4} filter and matched analysis/inverse reconstruction. The terminal low-pass is not treated as an additional directional scale.

\paragraph{Source-only directional statistics.}
All adaptation statistics are computed from source coefficients. For scale $s$ with $D_s$ native directions,
\begin{equation}
E_{s,d,t}=\operatorname{mean}_{c,h,w}\left|S^{\mathrm{src}}_{s,d,t}(c,h,w)\right|^2,
\qquad
q_{s,d,t}=\frac{E_{s,d,t}}{\sum_{j=1}^{D_s} E_{s,j,t}+\varepsilon}.
\label{eq:dir_energy}
\end{equation}
When the total energy is numerically zero, the implementation uses a uniform distribution. Equal directional energies are also mapped exactly to uniform weights. We define the mean-preserving relative directional importance
\begin{equation}
g_{s,d,t}=D_s q_{s,d,t}.
\label{eq:g}
\end{equation}
Thus $g_{s,d,t}=1$ under a directionally uniform source scale.

\paragraph{Scale scheduler.}
Before directional refinement, we compute a source-energy scale allocation. For the three-scale NSST, let the normalized scale energies, including the terminal low-pass, have coordinates $c=(0,1,\tfrac{1}{2},\tfrac{1}{4})$. Their energy centroid gives
\begin{equation}
\mathrm{centroid}_t=\sum_s \bar E_{s,t}c_s,
\qquad
\rho_t=\operatorname{clip}(1-\mathrm{centroid}_t,0,1),
\label{eq:centroid}
\end{equation}
followed by the complementary low-frequency scale weights
\begin{equation}
w^{\mathrm{LF}}_{0,t}=1,
\qquad
w^{\mathrm{LF}}_{s,t}=\frac{\rho_t^2}{\rho_t^2+c_s^2+\varepsilon}.
\label{eq:lf}
\end{equation}
The final configuration uses this LF branch; a separate HF fusion branch is disabled.

\paragraph{Threshold-free directional refinement.}
The Scale-only control broadcasts $w^{\mathrm{LF}}_{s,t}$ to every direction. Scale+Direction instead applies parameter-free odds modulation:
\begin{equation}
\boxed{
w_{s,d,t}
=
\frac{g_{s,d,t}w^{\mathrm{LF}}_{s,t}}
{g_{s,d,t}w^{\mathrm{LF}}_{s,t}+\left(1-w^{\mathrm{LF}}_{s,t}\right)}.}
\label{eq:dir_odds}
\end{equation}
There is no learned direction strength, frequency threshold, top-$k$, exponent, temperature, or direction-specific start/end schedule. Importantly, if $q_{s,d,t}=1/D_s$, then $g_{s,d,t}=1$ and Eq.~(\ref{eq:dir_odds}) reduces exactly to $w_{s,d,t}=w^{\mathrm{LF}}_{s,t}$. Directional adaptation therefore introduces no preference when the source scale itself is isotropic.

\subsection{Semantic Structure Mass-Odds: Where to Preserve}
\label{sec:ssmo}

The transform controller answers which source components are trusted, but not where they should act spatially. We derive a target-semantic map $A_t(p)\in[0,1]$ from the native target-conditioned attention at each step. For U-Net backbones, $A_t$ is the average cross-attention probability mass assigned to non-special target tokens across selected layers and heads. For SD3/FLUX.1-dev, we compute the analogous image-query probability mass over target-text tokens in native joint attention. Maps from different resolutions are aligned before aggregation. $A_t$ is probability mass, not a learned segmentation mask and not min--max normalized.

We combine target-semantic mass with the source prior using Semantic Structure Mass-Odds:
\begin{equation}
\boxed{
G_{s,d,t}(p)=
\frac{A_t(p)\,w_{s,d,t}}
{A_t(p)\,w_{s,d,t}+\left(1-w_{s,d,t}\right)}.}
\label{eq:ssmo}
\end{equation}
The implementation uses numerically stable endpoint handling so $w=0$ gives $G=0$ and $w=1$ gives $G=1$. A large $A_t(p)$ does not independently force source copying: it can increase the source share only when the source-derived prior $w_{s,d,t}$ is already non-negligible. We use target-token mass to identify locations participating in target-conditioned semantic structure, while the source controller determines whether the corresponding directional component is trusted. The directional coefficients are then fused pointwise,
\begin{equation}
\widetilde S_{s,d,t}(p)=
G_{s,d,t}(p)S^{\mathrm{src}}_{s,d,t}(p)
+\bigl(1-G_{s,d,t}(p)\bigr)S^{\mathrm{edit}}_{s,d,t}(p).
\label{eq:fusion}
\end{equation}
We deliberately keep the terminal low-pass from the edit branch,
\begin{equation}
\widetilde L_t=L^{\mathrm{edit}}_t,
\qquad
\widetilde z_{0,t}=\mathcal{T}^{-1}\!\left(\widetilde L_t,\{\widetilde S_{s,d,t}\}\right).
\label{eq:reconstruct}
\end{equation}
This yields the paper-wide interpretation
\begin{equation}
\underbrace{w_{s,d,t}}_{\text{WHAT + WHEN}}
\quad\times\quad
\underbrace{A_t(p)}_{\text{WHERE}}
\quad\longrightarrow\quad
\underbrace{G_{s,d,t}(p)}_{\text{source-preservation gate}}.
\end{equation}

\subsection{Task-Dependent Appearance Alignment}
\label{sec:appearance}

Structure preservation and appearance preservation require different controls. By default, for appearance-changing tasks, AdaIN and \dcrci are disabled so the requested color/style change is not pulled back toward the source. For appearance-preservation tasks, channel-wise AdaIN \citep{huang2017adain} first aligns the fused clean latent with the source clean prediction. \dcrci then performs a deterministic VAE decode--reencode cycle inside each guided step: it keeps the current lightness, inserts source CIELAB chroma at identity coordinates, and injects only the resulting latent cycle residual with unit weight. Its purpose is specifically to correct source-relative chroma drift, not to optimize every preservation metric. Appendix~\ref{app:algorithm} gives the exact construction, and Appendix~\ref{app:targeted_ablations} isolates its effect from AdaIN.

\subsection{Return to the Native Scheduler}
\label{sec:native_scheduler}

The final fused clean prediction is converted back to the update quantity expected by the installed scheduler: epsilon (or the configured equivalent) for DDIM and velocity for FlowMatch. The unmodified native scheduler then advances the edit state. Thus backbone-specific source construction, attention extraction, and scheduler semantics remain native, while the transform controller, \ssmo, reconstruction, and optional appearance alignment share one clean-prediction interface. Exact DDIM and FlowMatch conversions are given in Appendix~\ref{app:algorithm}.

\section{Experiments}
\label{sec:experiments}

\subsection{Experimental Setup}
\label{sec:setup}

\paragraph{Tasks and datasets.}
We evaluate two objectives. \textbf{Source-guided appearance changing} requires preserving geometry and identity while changing requested appearance; we use the official PIE-Bench Color40 subset ($n=40$) from PnP Inversion \citep{ju2024pnpinversion} and an ImageNet-R RGB-fixed variant ($n=87$ source--prompt pairs). The latter keeps the 29 ImageNet-R source images and first three official TI2I prompts fixed, then deterministically inserts red, blue, and green target colors into prompt indices 0, 1, and 2, respectively. It is a prompt-only derived subset, not an official ImageNet-R benchmark. \textbf{Source-guided appearance preservation} uses PIE-Bench Style-9 ($n=80$) and the first three official ImageNet-R-TI2I prompts for each of the same 29 mapped sources ($n=87$ pairs). ImageNet-R originates from \citet{hendrycks2021imagenetr}. All frozen manifests are method-independent.

\paragraph{Metrics.}
For both tasks, structural preservation is measured by DINO key self-similarity distance \citep{caron2021dino} and HED-edge similarity \citep{xie2015hed,radford2021clip}, while target adherence is measured by CLIP image--text cosine similarity (\CLIPT) \citep{radford2021clip}. For readability, \textbf{DINO distances are multiplied by $100$ in all reported tables}. Appearance-preservation tables additionally report CSD style similarity \citep{somepalli2024csd}, mean CIEDE2000 color difference $\Delta E_{00}$ \citep{sharma2005ciede2000}, and LPIPS-AlexNet \citep{zhang2018lpips}. CSD, $\Delta E_{00}$, and LPIPS are deliberately excluded from appearance-changing rankings because they reward retaining source appearance even when the prompt requests changing it. \CLIPT is a proxy for target-prompt alignment, and weak edits can score well on source-relative CSD.

\paragraph{Backbones and implementation.}
The primary benchmark uses SD2.1 at $512^2$ with 50 DDIM steps, CFG 7.5, and empty-text source inversion. Its U-Net-specific source self-attention Q/K injection (first 80\% of steps) and convolutional feature injection (all steps) are denoted jointly as \selfattnconv. The default NSST configuration uses three scales, Scale+Direction adaptation, \ssmo at every selected native step, LF scheduling, and HF fusion off. AdaIN/\dcrci are off for appearance changing and on with unit strength for appearance preservation. Cross-backbone experiments additionally use SD1.5, SD3-Medium at $1024^2$ with 50 native FlowMatch steps and CFG 7.0, and FLUX.1-dev at $1024^2$ with 28 native FlowMatch steps and guidance 3.5. SD3 and FLUX.1-dev use their native image-to-image strength contract and do not emulate \selfattnconv. All reported runs use seed 1. Experiments were executed on RTX 3090 GPUs; multi-GPU use only shards independent samples or configurations.

\paragraph{Baselines.}
We compare against PnP \citep{tumanyan2023pnp}, FSI-Edit \citep{yang2025fsiedit}, FreeDiff \citep{wu2024freediff}, FIA-Edit \citep{yang2026fiaedit}, Ctrl-X \citep{lin2024ctrlx}, FBSDiff \citep{gao2024fbsdiff}, FBSDiff++ \citep{gao2026fbsdiffpp}, DDPM-Inversion/Edit \citep{huberman2024ddpmedit}, and LEDITS++ \citep{brack2024ledits}. Competing methods use their official or locked local configurations; all generated outputs are then evaluated by one common paired evaluator. These comparisons therefore measure practical benchmark operating points, while the isolated contribution of direction adaptation is evaluated separately with matched ablations in Sec.~\ref{sec:directional_ablation}.

\subsection{Comparison with Existing Methods}
\label{sec:main_results}

Table~\ref{tab:appearance_change} evaluates appearance-changing edits. \method has the lowest DINO discrepancy and highest Edge similarity on both Color40 and ImageNet-R RGB-fixed, indicating stronger preservation of source geometry and boundaries. The trade-off is visible in \CLIPT: several baselines obtain higher target-text similarity, especially on ImageNet-R RGB-fixed. The result should therefore be read as a preservation--editability operating point rather than uniform dominance over every objective.

\begin{table}[t]
\caption{Source-guided appearance changing on PIE-Bench Color40 and ImageNet-R RGB-fixed. Lower DINO is better; higher Edge and \CLIPT are better. Bold denotes the best value within each dataset.}
\label{tab:appearance_change}
\centering
\tiny
\setlength{\tabcolsep}{2.7pt}
\resizebox{\linewidth}{!}{%
\begin{tabular}{lccc|ccc}
\toprule
& \multicolumn{3}{c|}{PIE-Bench Color40 ($n=40$)} & \multicolumn{3}{c}{ImageNet-R RGB-fixed ($n=87$)} \\
\cmidrule(lr){2-4}\cmidrule(lr){5-7}
Method & DINO $\downarrow$ & Edge $\uparrow$ & \CLIPT $\uparrow$ & DINO $\downarrow$ & Edge $\uparrow$ & \CLIPT $\uparrow$ \\
\midrule
PnP \citep{tumanyan2023pnp} & 0.3635 & 0.933883 & 0.320520 & 0.4692 & 0.923408 & 0.320213 \\
FSI-Edit \citep{yang2025fsiedit} & 0.2529 & 0.954036 & 0.308202 & 0.3047 & 0.946249 & 0.289174 \\
FreeDiff \citep{wu2024freediff} & 0.4014 & 0.912496 & 0.305592 & 0.5519 & 0.901492 & 0.300771 \\
FIA-Edit \citep{yang2026fiaedit} & 0.1776 & 0.958813 & 0.313540 & 0.2710 & 0.942602 & 0.299892 \\
Ctrl-X \citep{lin2024ctrlx} & 0.4487 & 0.926016 & 0.312394 & 0.4841 & 0.930784 & 0.296206 \\
FBSDiff \citep{gao2024fbsdiff} & 0.5948 & 0.909538 & \textbf{0.328082} & 0.6283 & 0.910846 & 0.317961 \\
FBSDiff++ \citep{gao2026fbsdiffpp} & 0.5828 & 0.920354 & 0.323535 & 0.6576 & 0.908733 & 0.322893 \\
DDPM-Inversion \citep{huberman2024ddpmedit} & 0.3012 & 0.940370 & 0.317586 & 0.4009 & 0.919397 & 0.316845 \\
LEDITS++ \citep{brack2024ledits} & 0.4208 & 0.914714 & 0.325675 & 0.6076 & 0.899558 & \textbf{0.333123} \\
\method (ours) & \textbf{0.1483} & \textbf{0.970599} & 0.310811 & \textbf{0.1977} & \textbf{0.965040} & 0.302801 \\
\bottomrule
\end{tabular}}
\end{table}

Table~\ref{tab:appearance_preserve} evaluates the harder preservation objective with six complementary metrics. On both Style-9 and ImageNet-R Official, \method obtains the strongest DINO, Edge, CSD, $\Delta E_{00}$, and LPIPS values among the evaluated methods. Its \CLIPT is lower than several baselines. This pattern is consistent across the two datasets: \method moves the operating point toward source structure, color, and perceptual fidelity while retaining non-trivial target alignment. The qualitative examples in Fig.~\ref{fig:qualitative} examine whether this preservation comes from visually ignoring the requested edit.

\begin{table}[t]
\caption{Source-guided appearance preservation. Higher Edge, \CLIPT, and CSD are better; lower DINO, $\Delta E_{00}$, and LPIPS are better. Bold denotes the best value in each panel.}
\label{tab:appearance_preserve}
\centering
\scriptsize
\textbf{(a) PIE-Bench Style-9 ($n=80$)}\\[-.25mm]
\resizebox{\linewidth}{!}{%
\begin{tabular}{lcccccc}
\toprule
Method & DINO $\downarrow$ & Edge $\uparrow$ & \CLIPT $\uparrow$ & CSD $\uparrow$ & $\Delta E_{00}\downarrow$ & LPIPS $\downarrow$ \\
\midrule
PnP \citep{tumanyan2023pnp} & 0.4650 & 0.931850 & \textbf{0.339857} & 0.585640 & 11.558568 & 0.364959 \\
FSI-Edit \citep{yang2025fsiedit} & 0.2739 & 0.953318 & 0.313164 & 0.785124 & 4.414129 & 0.190941 \\
FreeDiff \citep{wu2024freediff} & 0.4677 & 0.917802 & 0.313238 & 0.710960 & 8.046738 & 0.325889 \\
FIA-Edit \citep{yang2026fiaedit} & 0.2154 & 0.953988 & 0.323922 & 0.771782 & 4.867489 & 0.170106 \\
Ctrl-X \citep{lin2024ctrlx} & 0.4685 & 0.935669 & 0.314789 & 0.651415 & 12.728090 & 0.406545 \\
FBSDiff \citep{gao2024fbsdiff} & 0.5637 & 0.914867 & 0.327249 & 0.514609 & 24.627120 & 0.527509 \\
FBSDiff++ \citep{gao2026fbsdiffpp} & 0.5813 & 0.916741 & 0.328798 & 0.499506 & 25.948383 & 0.544150 \\
DDPM-Inversion \citep{huberman2024ddpmedit} & 0.3356 & 0.938427 & 0.325395 & 0.711654 & 7.714913 & 0.262373 \\
LEDITS++ \citep{brack2024ledits} & 0.4467 & 0.920429 & 0.328842 & 0.664181 & 10.832525 & 0.333747 \\
\method{} (ours) & \textbf{0.1535} & \textbf{0.971377} & 0.303636 & \textbf{0.830143} & \textbf{4.306591} & \textbf{0.161643} \\
\bottomrule
\end{tabular}}

\vspace{1.5mm}
\textbf{(b) ImageNet-R Official ($n=87$)}\\[-.25mm]
\resizebox{\linewidth}{!}{%
\begin{tabular}{lcccccc}
\toprule
Method & DINO $\downarrow$ & Edge $\uparrow$ & \CLIPT $\uparrow$ & CSD $\uparrow$ & $\Delta E_{00}\downarrow$ & LPIPS $\downarrow$ \\
\midrule
PnP \citep{tumanyan2023pnp} & 0.4711 & 0.926492 & 0.306057 & 0.481649 & 10.966128 & 0.366987 \\
FSI-Edit \citep{yang2025fsiedit} & 0.3028 & 0.945274 & 0.296027 & 0.677226 & 4.964035 & 0.214145 \\
FreeDiff \citep{wu2024freediff} & 0.5765 & 0.906282 & 0.308439 & 0.567379 & 9.305818 & 0.380600 \\
FIA-Edit \citep{yang2026fiaedit} & 0.2708 & 0.940538 & 0.302776 & 0.694631 & 5.842714 & 0.201335 \\
Ctrl-X \citep{lin2024ctrlx} & 0.4783 & 0.930707 & 0.291945 & 0.566347 & 12.892351 & 0.421791 \\
FBSDiff \citep{gao2024fbsdiff} & 0.6058 & 0.914581 & 0.311865 & 0.390922 & 28.879362 & 0.571614 \\
FBSDiff++ \citep{gao2026fbsdiffpp} & 0.6359 & 0.911147 & 0.313732 & 0.381843 & 30.209286 & 0.599476 \\
DDPM-Inversion \citep{huberman2024ddpmedit} & 0.3989 & 0.918887 & 0.313668 & 0.581791 & 9.179937 & 0.311198 \\
LEDITS++ \citep{brack2024ledits} & 0.5742 & 0.897639 & \textbf{0.317290} & 0.517529 & 13.180767 & 0.416248 \\
\method{} (ours) & \textbf{0.1744} & \textbf{0.968612} & 0.288178 & \textbf{0.753732} & \textbf{4.578329} & \textbf{0.192274} \\
\bottomrule
\end{tabular}}
\end{table}

\subsection{Qualitative Comparison}

\begin{figure}[t]
\centering
\includegraphics[width=\linewidth]{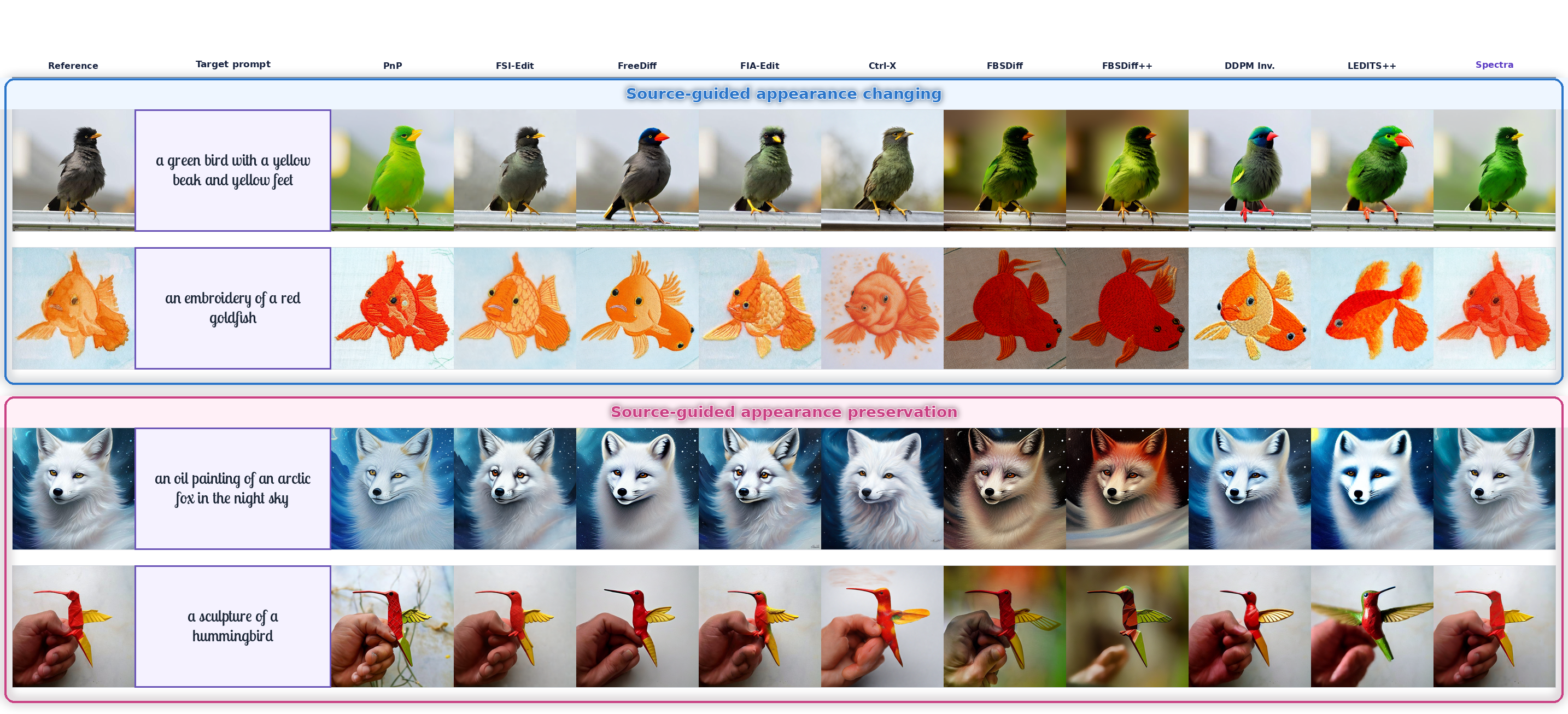}
\caption{Qualitative comparison against the evaluated baselines for source-guided appearance changing (top block) and appearance preservation (bottom block). The examples are selected to make both sides of the objective visible: the target attribute/style should change, while source geometry and non-target content should remain stable. Across these cases, \method exhibits less geometric drift and stronger source-local preservation; the comparison also makes clear that preservation should be judged jointly with whether the requested edit is actually expressed.}
\label{fig:qualitative}
\end{figure}

The appearance-changing rows in Fig.~\ref{fig:qualitative} illustrate the central failure mode behind the quantitative trade-off: methods can obtain stronger prompt alignment by changing broader image content than the requested attribute. Conversely, overly strong source locking can make the edit weak. In the appearance-preservation rows, \method retains local geometry and source appearance while changing the requested rendering style. We therefore use the full task-specific metric set rather than a single aggregate score.

\subsection{Transfer of the Preservation Interface Across Backbones}
\label{sec:backbones}

Table~\ref{tab:backbones} tests whether the clean-prediction preservation interface is tied to a U-Net diffusion architecture. SD1.5/SD2.1 use \selfattnconv in addition to the shared controller; SD3 and FLUX.1-dev instead use native joint attention and FlowMatch without emulating the U-Net injection path. Across all four backbones, \method moves the operating point toward lower DINO and higher Edge; for appearance preservation, CSD, $\Delta E_{00}$, and LPIPS also improve. At shared native img2img strength 0.4, \method lowers DINO by 20.7\% and 13.0\% on Color40 and by 35.9\% and 29.2\% on Style-9 for SD3-Medium and FLUX.1-dev, respectively. \CLIPT decreases in three of the four flow comparisons and increases slightly on Style-9 with FLUX.1-dev. These comparisons are not intended to attribute the full Vanilla-to-\method difference to the scale--direction controller: the U-Net variants additionally contain \selfattnconv, and the flow comparisons test the full pipelines at matched strength. Their narrower purpose is to establish that the same clean-prediction preservation interface can be instantiated across substantially different native pipelines without replacing their denoisers or schedulers. Controlled component effects are evaluated separately below.

\begin{table}[t]
\caption{Cross-backbone operating-point comparison on PIE-Bench. Bold marks the better Vanilla/\method value within each backbone pair and metric. SD3-Medium and FLUX.1-dev use native img2img strength 0.4 in both rows of each pair.}
\label{tab:backbones}
\centering
\tiny
\setlength{\tabcolsep}{2.0pt}
\resizebox{\linewidth}{!}{%
\begin{tabular}{lccc|cccccc}
\toprule
& \multicolumn{3}{c|}{Appearance changing: Color40 ($n=40$)} & \multicolumn{6}{c}{Appearance preservation: Style-9 ($n=80$)} \\
\cmidrule(lr){2-4}\cmidrule(lr){5-10}
Model & DINO $\downarrow$ & Edge $\uparrow$ & \CLIPT $\uparrow$ & DINO $\downarrow$ & Edge $\uparrow$ & \CLIPT $\uparrow$ & CSD $\uparrow$ & $\Delta E_{00}\downarrow$ & LPIPS $\downarrow$ \\
\midrule
Vanilla (SD1.5) & 0.7415 & 0.879225 & \textbf{0.326803} & 0.8310 & 0.884778 & \textbf{0.333407} & 0.477852 & 16.598175 & 0.542218 \\
\method (SD1.5) & \textbf{0.1321} & \textbf{0.971661} & 0.310350 & \textbf{0.1271} & \textbf{0.972858} & 0.298909 & \textbf{0.853918} & \textbf{4.319736} & \textbf{0.148353} \\
\midrule
Vanilla (SD2.1) & 0.7345 & 0.886537 & \textbf{0.333083} & 0.8463 & 0.879751 & \textbf{0.347094} & 0.450109 & 16.896389 & 0.548978 \\
\method (SD2.1) & \textbf{0.1483} & \textbf{0.970599} & 0.310811 & \textbf{0.1535} & \textbf{0.971377} & 0.303636 & \textbf{0.830143} & \textbf{4.306591} & \textbf{0.161643} \\
\midrule
Vanilla (SD3-Medium) & 0.23280 & 0.95803 & \textbf{0.30236} & 0.25694 & 0.95636 & \textbf{0.30829} & 0.80077 & 4.92691 & 0.22454 \\
\method (SD3-Medium) & \textbf{0.1845} & \textbf{0.963370} & 0.299517 & \textbf{0.1646} & \textbf{0.969802} & 0.296483 & \textbf{0.873107} & \textbf{3.054036} & \textbf{0.159466} \\
\midrule
Vanilla (FLUX.1-dev) & 0.21492 & 0.95865 & \textbf{0.29401} & 0.22416 & 0.95783 & 0.29171 & 0.88156 & 4.77811 & 0.22328 \\
\method (FLUX.1-dev) & \textbf{0.1870} & \textbf{0.965861} & 0.293674 & \textbf{0.1586} & \textbf{0.970501} & \textbf{0.292088} & \textbf{0.923750} & \textbf{3.243154} & \textbf{0.170879} \\
\bottomrule
\end{tabular}}
\end{table}

\subsection{Is Preservation Complementary Across Scales?}
\label{sec:scale_depth}

If preservation evidence is genuinely distributed across scales, progressively exposing additional transform scales should improve source retention with the same directional adaptation rule enabled at each depth. Table~\ref{tab:scale_depth} varies the complete NSST decomposition depth under the final preservation setting. Increasing $J$ from 1 to 3 monotonically improves DINO, Edge, CSD, $\Delta E_{00}$, and LPIPS, while \CLIPT changes modestly in the opposite direction and is highest at $J=1$. The ablation therefore supports cross-scale complementarity while making the operating-point trade-off explicit: additional scales move the result toward source fidelity at a small target-alignment cost.

\begin{table}[t]
\caption{NSST scale-depth ablation on PIE-Bench Style-9 ($n=80$), using the final SD2.1 preservation setting. $J$ is cumulative transform depth; direction adaptation remains enabled in every row.}
\label{tab:scale_depth}
\centering
\scriptsize
\resizebox{\linewidth}{!}{%
\begin{tabular}{clcccccc}
\toprule
$J$ & Included directional scales & DINO $\downarrow$ & Edge $\uparrow$ & \CLIPT $\uparrow$ & CSD $\uparrow$ & $\Delta E_{00}\downarrow$ & LPIPS $\downarrow$ \\
\midrule
1 & \{Scale 1\} & 0.2029 & 0.967400 & \textbf{0.310059} & 0.783478 & 5.067538 & 0.196868 \\
2 & \{Scale 1, Scale 2\} & 0.1802 & 0.970127 & 0.307306 & 0.799610 & 4.826044 & 0.180084 \\
3 & \{Scale 1, Scale 2, Scale 3\} & \textbf{0.1535} & \textbf{0.971377} & 0.303636 & \textbf{0.830143} & \textbf{4.306591} & \textbf{0.161643} \\
\bottomrule
\end{tabular}}
\end{table}

\subsection{Does Direction-Aware Allocation Help Beyond Scale Adaptation?}
\label{sec:directional_ablation}

Cross-scale heterogeneity alone does not establish that directions within the same scale should be treated differently. We therefore isolate within-scale direction adaptation with \selfattnconv, AdaIN, and \dcrci disabled while \ssmo and the LF scheduler remain fixed. Scale+Direction reduces DINO for SD1.5 ($0.6084\!\rightarrow\!0.5869$), SD2.1 ($0.6413\!\rightarrow\!0.6274$), SD3 ($0.3805\!\rightarrow\!0.3710$), and FLUX.1-dev ($0.3072\!\rightarrow\!0.3048$), while Edge, CSD, and \CLIPT are more mixed. Thus direction-aware allocation is a conditional refinement rather than a universal improvement knob. A stricter assignment control in Appendix~\ref{app:targeted_ablations} preserves the same non-uniform direction-weight multiset but cyclically misassigns it to different bands; the correct assignment gives the best DINO and Edge preservation on Color40. The complete cross-architecture metric table is reported in Appendix~\ref{app:direction_full}.

The Color40 paired bootstrap (10,000 resamples) gives DINO and Edge differences of $-0.001324$ and $+0.000656$ for correct versus shuffled assignment, with both 95\% intervals excluding zero. \CLIPT decreases by 0.000576. All three intervals for correct assignment versus Scale-only span zero (Table~\ref{tab:direction_uncertainty}). The depth ablation leaves the effect of adaptive scale weighting at fixed depth open.

\subsection{Mechanistic Analysis: When Does Direction Adaptation Help?}
\label{sec:mechanism}

If directional adaptation responds to a meaningful source property rather than merely adding flexibility, its advantage should be larger when the source itself is more directionally anisotropic. Ablation averages do not establish this link, so we test three questions under the same matched SD2.1 NSST isolation setting. For scale $s$ and timestep $t$, directional anisotropy is
\begin{equation}
\mathcal{A}_{s,t}=1-\frac{-\sum_{d=1}^{D_s}q_{s,d,t}\log q_{s,d,t}}{\log D_s}\in[0,1],
\label{eq:anisotropy}
\end{equation}
where zero denotes a uniform within-scale directional-energy distribution.

\begin{figure}[t]
    \centering
    \begin{minipage}[c]{0.45\linewidth}
        \centering

        \begin{subfigure}{\linewidth}
            \centering
            \includegraphics[
                width=\linewidth,
                height=3cm,
                keepaspectratio
            ]{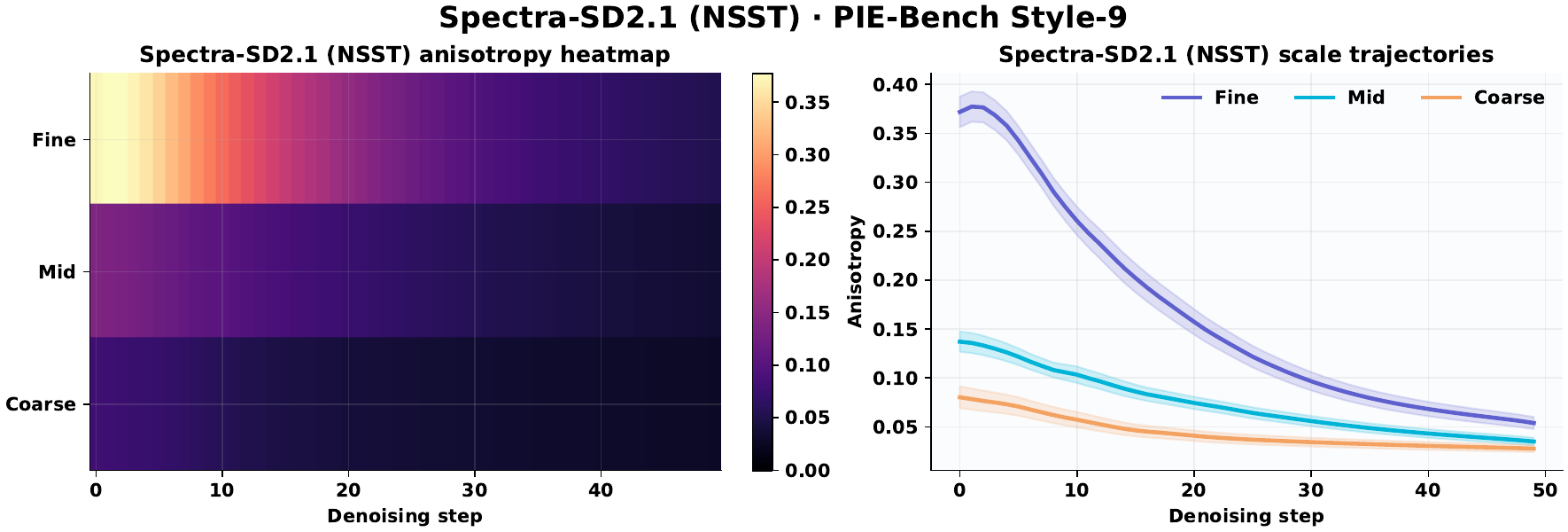}
        \caption{Scale- and time-dependent source anisotropy.}
            \label{fig:plot_a}
        \end{subfigure}

        \par\vspace{0.5em}

        \begin{subfigure}{\linewidth}
            \centering
            \includegraphics[
                width=\linewidth,
                height=3cm,
                keepaspectratio
            ]{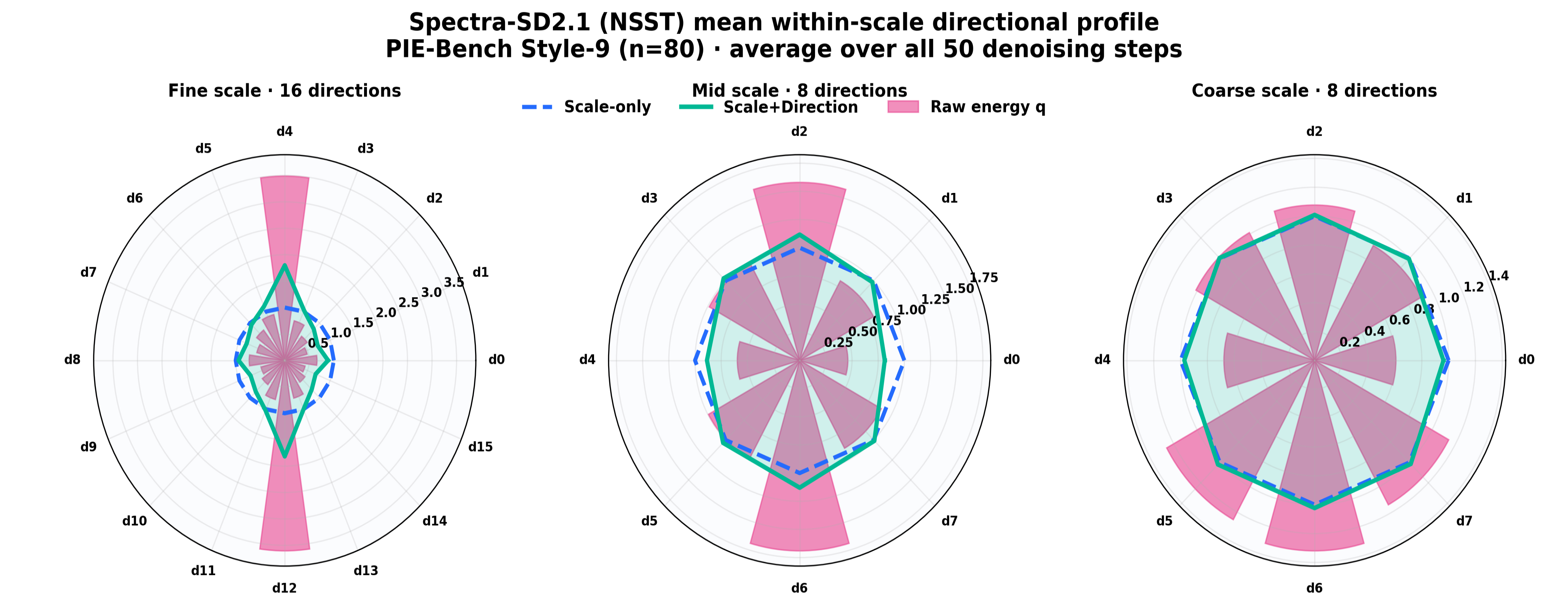}
        \caption{Realized within-scale directional treatment.}
            \label{fig:plot_b}
        \end{subfigure}
    \end{minipage}
    \hfill
    \begin{subfigure}[c]{0.5\linewidth}
        \centering
        \includegraphics[
            width=\linewidth,
            height=6.8cm,
            keepaspectratio
        ]{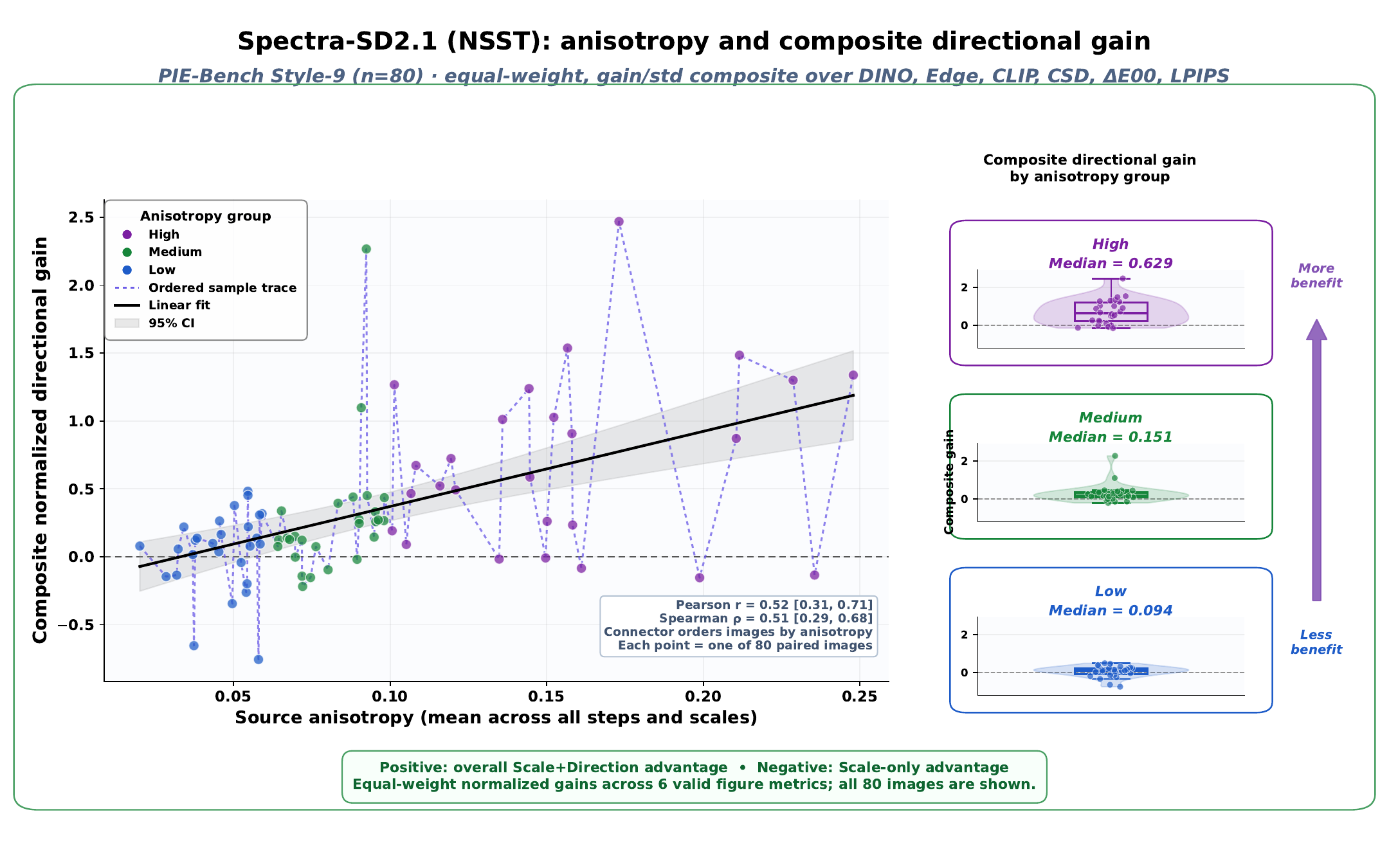}
        \caption{Source anisotropy versus paired Scale+Direction gain.}
        \label{fig:plot_c}
    \end{subfigure}
    \caption{Directional-heterogeneity mechanism audit for \method (SD2.1, NSST) on Style-9. (a) Source anisotropy varies across scale and denoising time. (b) Scale+Direction responds with non-uniform within-scale allocation, unlike Scale-only. (c) Larger source anisotropy is associated with larger paired directional gain; the reported correlation is descriptive, not causal.}
    \label{fig:mechanism_main}
\end{figure}

Figure~\ref{fig:mechanism_main} audits the mechanism. Panel (a) establishes a scale-dependent, time-varying anisotropy signal; panel (b) verifies that Scale+Direction, unlike the uniform Scale-only control, tracks the source directional profile. Panel (c) relates per-image mean anisotropy to an equal-weight composite of signed, standardized gains over DINO, Edge, \CLIPT, CSD, $\Delta E_{00}$, and LPIPS. The positive Pearson/Spearman association indicates that more anisotropic sources tend to obtain larger paired directional benefit; we do not interpret it causally.


\paragraph{Limitations.}\label{sec:limitations} Directional refinement is modest when source anisotropy is weak, and stronger preservation can reduce \CLIPT; \method therefore structures rather than removes the preservation--editability trade-off. The isolated \dcrci ablation likewise shows targeted gains in color/style fidelity rather than uniform improvement across all metrics, and appearance-preservation mode adds a VAE decode--reencode cycle. Finally, our experiments do not establish that the same policy is optimal when large non-rigid geometry changes require source structure itself to be discarded. Our evaluation focuses on color and rendering-style edits.

\subsection{Contribution of the Full Controller}
\label{sec:controlled_validation}

On SD2.1, a matched $2\times2$ ablation independently toggles \selfattnconv and the spectral controller, which combines LF fusion, adaptive scale--direction weighting, \ssmo, and reconstruction. Source construction, sampling, and each task's appearance settings are fixed. With injection enabled, adding the controller reduces mean DINO by 25.6\% on Color40 and 31.7\% on Style-9. Edge improves by 0.00741 and 0.00708, and \CLIPT declines by 0.00608 and 0.00811, respectively. DINO and Edge also improve with injection disabled (Appendix~\ref{app:factorial}).

\paragraph{Spatial and temporal allocation.}
On both SD2.1 datasets, \ssmo gives slightly better mean DINO and Edge than either a mean-matched constant gate or a spatially shuffled gate, with similar \CLIPT. Retaining per-step scale--direction weights lowers mean DINO by 0.00132 on Color40 and 0.00149 on Style-9 compared with using their temporal average with live target attention (Appendix~\ref{app:gate_controls}).

\subsection{Preservation--Editing Trade-off}
\label{sec:tradeoff}

We compare FIA-Edit with several SD2.1 \method configurations on the same image--prompt pairs for each task (Table~\ref{tab:tradeoff}). Full \method gives the best mean DINO and Edge on both datasets. Reduced injection raises mean \CLIPT on both tasks. On Style-9, disabling appearance alignment also raises \CLIPT, and Scale-only remains close to the default. For stronger target alignment, we recommend one NSST scale on Color40 and one scale with AdaIN enabled and \dcrci disabled on Style-9. These settings have the highest mean \CLIPT among the configurations shown and give higher Edge than FIA-Edit. Their DINO distance is higher on Color40 and nearly identical on Style-9. Users can select these settings through existing configuration parameters. Comparisons at equal visible edit-success rates remain open.

\begin{table}[tb]
\caption{PIE-Bench preservation--editing trade-off. Full \method uses three NSST scales. On Style-9, the balanced setting uses one NSST scale with AdaIN enabled and \dcrci disabled. Bold marks the best mean per dataset and metric.}
\label{tab:tradeoff}
\centering
\small
\setlength{\tabcolsep}{4pt}
\begin{tabular}{@{}lrrr rrr@{}}
\toprule
& \multicolumn{3}{c}{Color40 ($n=40$)} & \multicolumn{3}{c}{Style-9 ($n=80$)} \\
\cmidrule(lr){2-4}\cmidrule(lr){5-7}
Configuration & DINO $\downarrow$ & Edge $\uparrow$ & \CLIPT $\uparrow$ & DINO $\downarrow$ & Edge $\uparrow$ & \CLIPT $\uparrow$ \\
\midrule
FIA-Edit \citep{yang2026fiaedit} & 0.17761 & 0.95881 & 0.31354 & 0.21536 & 0.95399 & 0.32392 \\
Full \method & \textbf{0.14832} & \textbf{0.97060} & 0.31081 & \textbf{0.15350} & \textbf{0.97138} & 0.30364 \\
\method, reduced injection & 0.24655 & 0.94781 & 0.31410 & 0.24317 & 0.95924 & 0.31023 \\
\method, Scale-only & 0.1485 & 0.970097 & 0.310990 & 0.15772 & 0.97052 & 0.30381 \\
\method, appearance alignment off & --- & --- & --- & 0.16269 & 0.96423 & 0.31407 \\
\method, $J=1$ (balanced) & 0.18684 & 0.96602 & \textbf{0.31661} & 0.21582 & 0.96152 & \textbf{0.32506} \\
\bottomrule
\end{tabular}
\end{table}

\paragraph{Blinded human evaluation.}
Eighteen participants judged 20 uniformly sampled Style-9 cases, yielding 360 paired judgments of whether each output both applied the requested style and preserved shapes, poses, and layout. Balanced \method met both criteria in 135/360 judgments (37.5\%), and FIA-Edit in 112/360 (31.1\%). The paired difference was $+6.4$ percentage points (95\% CI $[-14.7,+27.5]$) using 10,000 two-way bootstrap resamples over cases and participants. The interval spans zero. Similar mean DINO on the full 80-case benchmark motivated the comparison. Appendix~\ref{app:human_eval} gives the protocol and full results.

\paragraph{Computational cost.}
On an RTX 3090 at batch size 1, \method adds 14.07--25.61~s in first-edit latency and 0.02--1.39~GiB in peak memory over Vanilla on Color40, and 26.88--43.87~s and 0.73--1.40~GiB on Style-9 across SD2.1 and SD3-Medium. Disabling \dcrci saves 13.00~s on SD2.1 and 17.95~s on SD3-Medium on Style-9. Appendix~\ref{app:runtime} gives the full profiles.

\section{Conclusion}
\label{sec:conclusion}

We studied source preservation in text-guided image editing as a structured allocation problem rather than a fixed frequency-selection problem. The evidence reveals a hierarchy: increasing transform depth improves source retention, while the smaller benefit of within-scale direction adaptation is associated with source anisotropy. \method converts these source statistics into threshold-free scale--direction weights, couples them with target-semantic spatial evidence through \ssmo, and returns the fused clean prediction to the backbone's native scheduler. The shuffled-direction control further shows better structural preservation when weights are assigned to their corresponding directional bands, while the isolated \dcrci study confirms its narrower role in shifting appearance-preservation outputs toward source color/style fidelity. Across diffusion and flow backbones, these results support treating source preservation as a dynamic Scale $\times$ Direction $\times$ Space $\times$ Time allocation problem rather than assigning fixed semantic roles to frequency bands.

\clearpage
\section*{Reproducibility Statement}
\label{sec:repro_statement}
The submission is accompanied by code and frozen experiment manifests. Section~\ref{sec:setup} states the main evaluation protocol and task-dependent metrics; Appendix~\ref{app:datasets} specifies the benchmark subsets (Table~\ref{tab:dataset_subsets}) and prompt construction; Appendix~\ref{app:backbones} records backbone-specific generation settings; Appendix~\ref{app:metrics} gives metric definitions and preprocessing; and Appendix~\ref{app:algorithm} provides the complete per-step pseudocode and scheduler conversions. Within each controlled model comparison, source IDs, target prompts, and the common evaluator are fixed. Generation settings are shared except for the intervention being tested, and incomplete sample sets are rejected rather than silently averaged. Appendices~\ref{app:targeted_ablations}, \ref{app:runtime}, \ref{app:controlled_sd}, and~\ref{app:human_eval} document the additional protocols.

\section*{AI Use Statement}
Generative AI tools were used to assist with literature search and summarization, methodological and experimental framing, code debugging, organization and interpretation of already-computed results, figure/table presentation, and manuscript drafting and language editing. All experimental outputs, numerical results, dataset manifests, and implementation artifacts reported in the paper were produced and checked by the authors. The authors manually reviewed the cited literature and verified the final claims, equations, references, tables, figures, and AI-assisted text, and take responsibility for the complete content of the submission. For this revision, generative AI also assisted with organizing the completed rebuttal evidence, drafting edits, and implementing them in LaTeX.

\section*{Ethics Statement}
The Style-9 study collected paired judgments of editing outputs with method identities concealed. We report aggregate responses and confidence intervals (Appendix~\ref{app:human_eval}).
Ethics approval was not required for the blinded Style-9 human evaluation.

\bibliography{iclr2027_conference,spectra_wedit}
\bibliographystyle{iclr2027_conference}

\clearpage
\appendix
\addtocounter{table}{-1}
\renewcommand{\theHtable}{appendix.\Alph{section}.\arabic{table}}

\section{Complete SPECTRA Procedure}
\label{app:algorithm}

\subsection{Per-step pseudocode}
\begin{algorithm}[t]
\caption{One native \method sampling step at timestep $t$.}
\label{alg:spectra}
\begin{algorithmic}[1]
\Require source state $z_t^{\mathrm{src}}$, edit state $z_t^{\mathrm{edit}}$, target prompt $p^{\mathrm{tgt}}$
\State predict $\hat z^{\mathrm{src}}_{0,t}$ under empty text and $\hat z^{\mathrm{edit}}_{0,t}$ under $p^{\mathrm{tgt}}$
\State record target-semantic probability mass $A_t$ from native cross/joint attention
\State $(L_t^{\mathrm{src}},\{S^{\mathrm{src}}_{s,d,t}\})\gets\mathcal T(\hat z^{\mathrm{src}}_{0,t})$; $(L_t^{\mathrm{edit}},\{S^{\mathrm{edit}}_{s,d,t}\})\gets\mathcal T(\hat z^{\mathrm{edit}}_{0,t})$
\State compute source scale energies, $w^{\mathrm{LF}}_{s,t}$, directional energies $E_{s,d,t}$, and $w_{s,d,t}$
\State $G_{s,d,t}\gets\operatorname{SSMO}(A_t,w_{s,d,t})$ and fuse directional coefficients by Eq.~(\ref{eq:fusion})
\State set $\widetilde L_t\gets L_t^{\mathrm{edit}}$ and reconstruct $\widetilde z_{0,t}\gets\mathcal T^{-1}(\widetilde L_t,\{\widetilde S_{s,d,t}\})$
\If{appearance alignment is enabled}
    \State apply AdaIN, followed by \dcrci if enabled, to obtain $\hat z^{\mathrm{merged}}_{0,t}$
\Else
    \State $\hat z^{\mathrm{merged}}_{0,t}\gets\widetilde z_{0,t}$
\EndIf
\State convert $\hat z^{\mathrm{merged}}_{0,t}$ to the native DDIM noise or FlowMatch velocity
\State execute exactly one unmodified native scheduler step to obtain $z_{t-1}^{\mathrm{edit}}$
\end{algorithmic}
\end{algorithm}
Algorithm~\ref{alg:spectra} makes the separation explicit: backbone-specific prediction, attention capture, and scheduler conversion stay native, whereas the transform controller, \ssmo fusion, and optional appearance alignment act on clean latent predictions. No \method component changes the backbone weights.

\subsection{Diffusion source trajectory and DDIM conversion}
SD1.5 and SD2.1 encode the source image with the frozen VAE and construct a deterministic empty-text DDIM inversion trajectory using 999 inversion steps. The latent states corresponding to the 50 generation steps are cached. For epsilon prediction, the usual clean estimate is
\begin{equation}
\hat z_0=\frac{z_t-\sqrt{1-\bar\alpha_t}\,\epsilon_t}{\sqrt{\bar\alpha_t}}.
\end{equation}
After \method fusion, the epsilon implied by the desired clean prediction is recomputed as
\begin{equation}
\epsilon_t^{\mathrm{merged}}=
\frac{z_t^{\mathrm{edit}}-\sqrt{\bar\alpha_t}\,\hat z_{0,t}^{\mathrm{merged}}}{\sqrt{1-\bar\alpha_t}},
\label{eq:ddim_back}
\end{equation}
and passed to the unmodified scheduler. The reported DDIM configuration is deterministic ($\eta=0$).

\subsection{Flow source state and native conversion}
For SD3 and FLUX.1-dev, a single seed-fixed noise realization $\epsilon$ is shared by source-state reconstruction and target initialization. At a selected native FlowMatch step with scheduler coordinate $\sigma_t$, the source state is generated through the installed scheduler's native noising function; under the standard linear interpretation,
\begin{equation}
z_t^{\mathrm{src}}=(1-\sigma_t)z_0^{\mathrm{src}}+\sigma_t\epsilon.
\end{equation}
A predicted velocity $v_t$ corresponds to a clean prediction $\hat z_0=z_t-\sigma_t v_t$. After fusion, the compatible velocity is
\begin{equation}
 v_t^{\mathrm{merged}}=\frac{z_t^{\mathrm{edit}}-\hat z_{0,t}^{\mathrm{merged}}}{\sigma_t},
\label{eq:flow_back}
\end{equation}
and the installed FlowMatch Euler scheduler advances the sample. The strength parameter selects a suffix of the official native image-to-image schedule; \method uses the sigma aligned with that selected step rather than the first sigma of the untruncated schedule.

\subsection{Native semantic probability mass}
For Stable Diffusion, let $P_t^{(\ell,h)}(p,k)$ be cross-attention probability from spatial position $p$ to token $k$ at layer $\ell$ and head $h$. With $\mathcal{T}_{\mathrm{text}}$ the non-special target-token set,
\begin{equation}
A_t(p)=\frac{1}{|\mathcal L||\mathcal H|}
\sum_{\ell\in\mathcal L}\sum_{h\in\mathcal H}\sum_{k\in\mathcal T_{\mathrm{text}}}
P_t^{(\ell,h)}(p,k).
\end{equation}
Maps from different native resolutions are bilinearly aligned before layer aggregation. For SD3/FLUX.1-dev, the recorder uses image queries over the full native joint key sequence, sums attention probability assigned to target-text positions, then averages over heads and eligible blocks. The recorder is active only during the target-conditioned edit pass.

\subsection{DCRCI details}
Let $x^{\mathrm{base}}$ be the RGB image decoded from the AdaIN-aligned latent and let $(L^{\mathrm{base}},a^{\mathrm{base}},b^{\mathrm{base}})$ be its D65 CIELAB representation. Let $(a^{\mathrm{src}},b^{\mathrm{src}})$ be source chroma at the same image coordinates. \method forms a gamut-safe color target
\begin{equation}
x^{\mathrm{color}}=\operatorname{Lab}^{-1}_{\mathrm{gamut-safe}}
\left(L^{\mathrm{base}},a^{\mathrm{src}},b^{\mathrm{src}}\right),
\end{equation}
using 12 deterministic gamut-mapping iterations. The active frozen VAE deterministically encodes $x^{\mathrm{base}}$ and $x^{\mathrm{color}}$ to $z^{\mathrm{base}}$ and $z^{\mathrm{color}}$, then injects only the cycle residual,
\begin{equation}
\hat z^{\mathrm{merged}}_{0,t}=z^{\mathrm{AdaIN}}_{0,t}+\lambda_C\left(z^{\mathrm{color}}_{0,t}-z^{\mathrm{base}}_{0,t}\right),\qquad \lambda_C=1.
\label{eq:dcrci}
\end{equation}
Thus \dcrci preserves the current lightness and geometry while correcting source-relative chroma drift.

\section{Datasets and Benchmark Governance}
\label{app:datasets}

\begin{table}[h]
\renewcommand{\thetable}{B.1}
\caption{Frozen task-specific evaluation subsets. ``Cases'' denotes source--prompt pairs; ImageNet-R Official and RGB-fixed share the same 29 source images and differ only in target text.}
\label{tab:dataset_subsets}
\centering
\scriptsize
\resizebox{\linewidth}{!}{%
\begin{tabular}{lllll}
\toprule
Family & Subset & Task & Cases / sources & Prompt protocol \\
\midrule
PIE-Bench & Color40 & Appearance changing & 40 / 40 & Official \texttt{editing\_prompt} \\
PIE-Bench & Style-9 & Appearance preservation & 80 / 80 & Official \texttt{editing\_prompt} \\
ImageNet-R-TI2I & Official & Appearance preservation & 87 / 29 & First three official target prompts \\
ImageNet-R-TI2I & RGB-fixed & Appearance changing & 87 / 29 & Same prompts with deterministic red/blue/green class-color insertion \\
\bottomrule
\end{tabular}}
\end{table}

PIE-Bench contains 700 annotated images over ten editing types \citep{ju2024pnpinversion}. We use its official \texttt{6\_change\_attribute\_color\_40} and \texttt{9\_change\_style\_80} subsets. Square-bracket annotation markers are removed without changing semantic words. The ImageNet-R-TI2I mapping distributed with PIE-Bench/PnP Inversion supplies five target prompts for each of 29 mapped ImageNet-R sources; we retain the first three. For RGB-fixed, if $p_{i,k}$ is the $k$-th retained official target prompt and $c_i$ the class noun, we construct
\begin{equation}
\tilde p_{i,k}=\operatorname{insert}(p_{i,k},c_i,\gamma_k),
\qquad
\gamma_0=\texttt{red},\;\gamma_1=\texttt{blue},\;\gamma_2=\texttt{green}.
\end{equation}
The builder fails if the class noun is absent, and stores both the original and edited prompt. No image selection is based on model outputs. Every frozen manifest stores a relative source path, source identifier, target prompt, and verified image size; all source images used here are $512\times512$.

\section{Evaluation Metrics}
\label{app:metrics}

All source-output metrics use the same paired source $x^{\mathrm{src}}$, generated image $x^{\mathrm{gen}}$, and target prompt $p^{\mathrm{tgt}}$. If a method generates at a different native resolution, $x^{\mathrm{gen}}$ is bicubically resized to the source resolution only for the temporary source-output metric pair; the saved generation is not overwritten.

\paragraph{DINO structure distance.}
We use a pinned DINO ViT-B/8 \citep{caron2021dino}. From block 11 attention keys, including the CLS token, each token key is $\ell_2$-normalized and used to form a cosine self-similarity matrix. DINO is the MSE between source and output self-similarity matrices. All paper tables multiply the raw MSE by $100$.

\paragraph{Edge similarity.}
HED \citep{xie2015hed} produces edge maps for source and output. We embed the edge maps with the same CLIP image encoder \citep{radford2021clip} and report cosine similarity.

\paragraph{Target-text alignment.}
\CLIPT is the cosine similarity between the generated-image CLIP embedding and the embedding of $p^{\mathrm{tgt}}$. The source prompt is never used for this metric.

\paragraph{Appearance-preservation metrics.}
CSD uses the official ViT-L style descriptor and reports source-output cosine similarity \citep{somepalli2024csd}. $\Delta E_{00}$ is mean pixel-wise CIEDE2000 distance after a common sRGB-to-CIELab D65 conversion with unit parametric factors \citep{sharma2005ciede2000}. LPIPS uses the official AlexNet v0.1 model \citep{zhang2018lpips}. These three metrics are reported only when retaining source appearance is part of the task objective.

\section{Backbone and Task Configuration}
\label{app:backbones}

\begin{table}[h]
\caption{Backbone-specific generation settings used by \method. For the main flow comparison in Table~\ref{tab:backbones}, \method and native img2img both use strength 0.4. Appendix~\ref{app:strength} reports the \method strength sweep and earlier native 0.6 baselines (Table~\ref{tab:native_strength_orig}).}
\centering
\scriptsize
\resizebox{\linewidth}{!}{%
\begin{tabular}{lccccp{5.1cm}}
\toprule
Backbone & Resolution & Native steps & Guidance & Strength & Source-state construction \\
\midrule
SD1.5 & $512^2$ & 50 DDIM & CFG 7.5 & -- & Empty-text DDIM inversion; 999 inversion steps; matched source states cached \\
SD2.1 & $512^2$ & 50 DDIM & CFG 7.5 & -- & Empty-text DDIM inversion; 999 inversion steps; matched source states cached \\
SD3-Medium & $1024^2$ & 50 FlowMatch & CFG 7.0 & 0.4 & Native img2img schedule, seed-fixed flow noise, scheduler-native source noising \\
FLUX.1-dev & $1024^2$ & 28 FlowMatch & guidance 3.5 & 0.4 & Native img2img schedule, seed-fixed flow noise, scheduler-native source noising \\
\bottomrule
\end{tabular}}
\end{table}

For SD1.5/SD2.1, \selfattnconv denotes source self-attention Q/K injection over the first 80\% of generation steps together with source convolutional feature injection over all steps. This U-Net-specific path is not emulated for SD3 or FLUX.1-dev. Across all backbones, the default \method controller uses three-scale NSST, Scale+Direction adaptation, \ssmo at every selected native step, LF scheduling on, HF fusion off, and empty-text source conditioning. AdaIN and \dcrci are off for Color40/RGB-fixed and on with unit strength for Style-9/ImageNet-R Official.

Table~\ref{tab:tradeoff} documents the balanced operating points. For SD2.1, full \method uses attention and convolutional injection windows $(a_t,f_t)=(0.8,1.0)$, and the reduced setting uses $(0.5,0.8)$. On Style-9, the appearance-alignment-off setting disables AdaIN and \dcrci, and the balanced setting enables only AdaIN. Dashes in the table denote an unreported configuration for that dataset.

The directional-heterogeneity isolation studies intentionally differ from this final configuration: \selfattnconv, AdaIN, \dcrci, and HF fusion are off; \ssmo and LF scheduling stay on. This distinction prevents appearance alignment or U-Net-specific injection from being attributed to the directional controller.

\section{SD3/FLUX.1-dev Strength Sensitivity}
\label{app:strength}

The native image-to-image strength changes the amount of the native schedule traversed and therefore moves the preservation--editability operating point. Tables~\ref{tab:strength_change} and \ref{tab:strength_preserve} sweep only this parameter.

\begin{table}[h]
\caption{Strength sensitivity for appearance changing on PIE-Bench Color40 ($n=40$). \ssmo and LF scheduling are on; \selfattnconv, AdaIN, and \dcrci are off.}
\label{tab:strength_change}
\centering
\scriptsize
\begin{tabular}{lcccc}
\toprule
Model & Strength & DINO $\downarrow$ & Edge $\uparrow$ & \CLIPT $\uparrow$ \\
\midrule
SD3 & 0.4 & \textbf{0.1845} & \textbf{0.963370} & 0.299517 \\
SD3 & 0.5 & 0.2606 & 0.953579 & 0.302705 \\
SD3 & 0.6 & 0.3555 & 0.943482 & \textbf{0.303805} \\
\midrule
FLUX.1-dev & 0.4 & \textbf{0.1870} & \textbf{0.965861} & 0.293674 \\
FLUX.1-dev & 0.5 & 0.2418 & 0.954120 & 0.292217 \\
FLUX.1-dev & 0.6 & 0.2858 & 0.944901 & \textbf{0.295297} \\
\bottomrule
\end{tabular}
\end{table}

\begin{table}[h]
\caption{Strength sensitivity for appearance preservation on PIE-Bench Style-9 ($n=80$). AdaIN and \dcrci are enabled; \selfattnconv is off.}
\label{tab:strength_preserve}
\centering
\scriptsize
\resizebox{\linewidth}{!}{%
\begin{tabular}{llcccccc}
\toprule
Model & Strength & DINO $\downarrow$ & Edge $\uparrow$ & \CLIPT $\uparrow$ & CSD $\uparrow$ & $\Delta E_{00}\downarrow$ & LPIPS $\downarrow$ \\
\midrule
SD3 & 0.4 & \textbf{0.1646} & \textbf{0.969802} & 0.296483 & \textbf{0.873107} & \textbf{3.054036} & \textbf{0.159466} \\
SD3 & 0.5 & 0.2233 & 0.958931 & 0.302507 & 0.830049 & 3.619108 & 0.203665 \\
SD3 & 0.6 & 0.2961 & 0.949617 & \textbf{0.309107} & 0.782773 & 4.329398 & 0.249465 \\
\midrule
FLUX.1-dev & 0.4 & \textbf{0.1586} & \textbf{0.970501} & 0.292088 & \textbf{0.923750} & \textbf{3.243154} & \textbf{0.170879} \\
FLUX.1-dev & 0.5 & 0.2181 & 0.959972 & 0.294987 & 0.893549 & 3.867091 & 0.222740 \\
FLUX.1-dev & 0.6 & 0.2610 & 0.953944 & \textbf{0.295888} & 0.877777 & 4.392754 & 0.258915 \\
\bottomrule
\end{tabular}}
\end{table}

The monotonic preservation trend and the weaker/non-monotonic \CLIPT trend are why we treat 0.4 as a selected preservation-oriented operating point rather than a universally optimal strength.

Table~\ref{tab:native_strength_orig} records the original native img2img results at strength 0.6. Table~\ref{tab:backbones} reports the flow comparisons at shared strength 0.4.

\begin{table}[H]
\addtocounter{table}{-1}
\renewcommand{\thetable}{E.1}
\renewcommand{\theHtable}{native-original}
\caption{Original native img2img baselines at strength 0.6 on PIE-Bench. Values are dataset means.}
\label{tab:native_strength_orig}
\centering
\footnotesize
\resizebox{\linewidth}{!}{%
\begin{tabular}{@{}lrrr|rrrrrr@{}}
\toprule
& \multicolumn{3}{c|}{Color40 ($n=40$)} & \multicolumn{6}{c}{Style-9 ($n=80$)} \\
\cmidrule(lr){2-4}\cmidrule(lr){5-10}
Backbone & DINO $\downarrow$ & Edge $\uparrow$ & \CLIPT $\uparrow$ & DINO $\downarrow$ & Edge $\uparrow$ & \CLIPT $\uparrow$ & CSD $\uparrow$ & $\Delta E_{00}\downarrow$ & LPIPS $\downarrow$ \\
\midrule
SD3-Medium & 0.4193 & 0.929038 & 0.308297 & 0.4594 & 0.928791 & 0.325140 & 0.673228 & 7.751022 & 0.366294 \\
FLUX.1-dev & 0.3263 & 0.935293 & 0.296042 & 0.3470 & 0.938004 & 0.295440 & 0.825216 & 6.758716 & 0.322700 \\
\bottomrule
\end{tabular}}
\end{table}

\section{Full Cross-Architecture Directional Ablation}
\label{app:direction_full}
The main text reports the DINO trend and the mechanistic analysis. Table~\ref{tab:direction_arch} gives the complete matched Scale-only/Scale+Direction comparison. \selfattnconv, AdaIN, and \dcrci are off; \ssmo and the LF scheduler are on, so the two rows within each backbone differ only in the direction-aware odds modulation.

\begin{table}[H]
\caption{Matched Scale-only versus Scale+Direction ablation on PIE-Bench Style-9 ($n=80$) with NSST. \ssmo and the LF scheduler are on; \selfattnconv, AdaIN, and \dcrci are off. Bold marks the better value within each backbone pair.}
\label{tab:direction_arch}
\centering
\scriptsize
\resizebox{\linewidth}{!}{%
\begin{tabular}{llccccc}
\toprule
Backbone & Adaptation & DINO $\downarrow$ & Edge $\uparrow$ & \CLIPT $\uparrow$ & CSD $\uparrow$ & $\Delta E_{00}\downarrow$ \\
\midrule
\multirow{2}{*}{SD1.5} & Scale-only & 0.6084 & 0.901989 & 0.329210 & 0.591019 & 12.561252 \\
 & Scale+Direction & \textbf{0.5869} & \textbf{0.906085} & \textbf{0.329934} & \textbf{0.596218} & \textbf{12.355466} \\
\midrule
\multirow{2}{*}{SD2.1} & Scale-only & 0.6413 & 0.902645 & \textbf{0.342469} & \textbf{0.558335} & 12.360673 \\
 & Scale+Direction & \textbf{0.6274} & \textbf{0.904151} & 0.341949 & 0.557640 & \textbf{12.097185} \\
\midrule
\multirow{2}{*}{SD3} & Scale-only & 0.3805 & \textbf{0.939675} & \textbf{0.317157} & 0.719125 & 6.073878 \\
 & Scale+Direction & \textbf{0.3710} & 0.939357 & 0.316005 & \textbf{0.724979} & \textbf{5.968601} \\
\midrule
\multirow{2}{*}{FLUX.1-dev} & Scale-only & 0.3072 & 0.945525 & \textbf{0.297217} & 0.849005 & 5.889809 \\
 & Scale+Direction & \textbf{0.3048} & \textbf{0.945923} & 0.297040 & \textbf{0.849686} & \textbf{5.849706} \\
\bottomrule
\end{tabular}}
\end{table}

\section{Transform-General Directional Ablations}
\label{app:transforms}

The default NSST is useful because it exposes multiple native directions at each scale, but the controller is defined over generic directional subbands. We therefore repeat the Scale-only/Scale+Direction isolation with DTCWT, DWT, and NSCT. Each transform keeps its native analysis/synthesis and number of directions; the scale scheduler, source-only energy normalization, odds rule, \ssmo gate, prompts, seeds, and evaluation protocol are fixed.

\begin{table}[H]
\caption{Transform-general directional-heterogeneity ablation on PIE-Bench Style-9 ($n=80$). \selfattnconv, AdaIN, and \dcrci are off; \ssmo and LF scheduling are on. Bold marks the better value within each backbone--transform Scale-only/Scale+Direction pair.}
\label{tab:transform_all}
\centering
\tiny
\resizebox{\linewidth}{!}{%
\begin{tabular}{lllccccc}
\toprule
Backbone & Transform & Adaptation & DINO $\downarrow$ & Edge $\uparrow$ & \CLIPT $\uparrow$ & CSD $\uparrow$ & $\Delta E_{00}\downarrow$ \\
\midrule
SD1.5 & DTCWT & Scale-only & 0.630900 & 0.900893 & \textbf{0.331566} & 0.570083 & 13.786419 \\
SD1.5 & DTCWT & Scale+Direction & \textbf{0.619300} & \textbf{0.903570} & 0.330081 & \textbf{0.574700} & \textbf{13.649394} \\
SD1.5 & DWT & Scale-only & 0.619900 & 0.904396 & \textbf{0.331286} & 0.571097 & 13.831777 \\
SD1.5 & DWT & Scale+Direction & \textbf{0.613300} & \textbf{0.905741} & 0.330079 & \textbf{0.574846} & \textbf{13.727056} \\
SD1.5 & NSCT & Scale-only & 0.623900 & \textbf{0.898967} & \textbf{0.330926} & 0.583754 & 12.009912 \\
SD1.5 & NSCT & Scale+Direction & \textbf{0.623600} & 0.898511 & 0.330653 & \textbf{0.585887} & \textbf{12.007414} \\
\midrule
SD2.1 & DTCWT & Scale-only & 0.667400 & 0.899972 & 0.343685 & 0.539182 & 13.564509 \\
SD2.1 & DTCWT & Scale+Direction & \textbf{0.653600} & \textbf{0.903597} & \textbf{0.344153} & \textbf{0.542858} & \textbf{13.387058} \\
SD2.1 & DWT & Scale-only & 0.651700 & 0.902969 & \textbf{0.342219} & 0.541375 & 13.661466 \\
SD2.1 & DWT & Scale+Direction & \textbf{0.649500} & \textbf{0.903659} & 0.342041 & \textbf{0.545559} & \textbf{13.518173} \\
SD2.1 & NSCT & Scale-only & 0.655900 & \textbf{0.898018} & \textbf{0.343129} & 0.555116 & 11.812372 \\
SD2.1 & NSCT & Scale+Direction & \textbf{0.655400} & 0.897570 & 0.343015 & \textbf{0.555544} & \textbf{11.810691} \\
\midrule
SD3 & DTCWT & Scale-only & 0.381400 & \textbf{0.939024} & \textbf{0.316698} & 0.721651 & 6.133696 \\
SD3 & DTCWT & Scale+Direction & \textbf{0.375000} & 0.938867 & 0.316216 & \textbf{0.724296} & \textbf{6.063314} \\
SD3 & DWT & Scale-only & 0.381700 & \textbf{0.940139} & \textbf{0.316670} & 0.723406 & 6.245104 \\
SD3 & DWT & Scale+Direction & \textbf{0.377000} & 0.939685 & 0.316208 & \textbf{0.725869} & \textbf{6.200093} \\
SD3 & NSCT & Scale-only & 0.387900 & 0.938528 & \textbf{0.316396} & 0.720356 & 5.918791 \\
SD3 & NSCT & Scale+Direction & \textbf{0.387400} & \textbf{0.938987} & 0.316342 & \textbf{0.720784} & \textbf{5.911797} \\
\midrule
FLUX.1-dev & DTCWT & Scale-only & 0.307800 & \textbf{0.945696} & \textbf{0.296936} & 0.848719 & 5.904492 \\
FLUX.1-dev & DTCWT & Scale+Direction & \textbf{0.305500} & 0.945530 & 0.296872 & \textbf{0.849322} & \textbf{5.865041} \\
FLUX.1-dev & DWT & Scale-only & 0.306200 & \textbf{0.945432} & 0.296956 & 0.849560 & 5.900141 \\
FLUX.1-dev & DWT & Scale+Direction & \textbf{0.304100} & 0.945084 & \textbf{0.297109} & \textbf{0.850499} & \textbf{5.886910} \\
FLUX.1-dev & NSCT & Scale-only & 0.310800 & 0.944567 & 0.297045 & 0.847266 & 5.921114 \\
FLUX.1-dev & NSCT & Scale+Direction & \textbf{0.310400} & \textbf{0.944809} & \textbf{0.297069} & \textbf{0.847692} & \textbf{5.917606} \\
\bottomrule
\end{tabular}}
\end{table}

For SD2.1, Tables~\ref{tab:sd21_pres_transform} and \ref{tab:sd21_change_transform} also include NSST. Table~\ref{tab:sd21_pres_transform} additionally reports LPIPS. The transform study does not establish that every transform benefits on every metric. Its narrower conclusion is that the source-only direction rule can be instantiated on different native directional decompositions without transform-specific tuning.

\begin{table}[H]
\caption{SD2.1 transform ablation for appearance preservation on Style-9 ($n=80$) under the directional isolation setting described in Appendix~\ref{app:backbones}.}
\label{tab:sd21_pres_transform}
\centering
\scriptsize
\resizebox{\linewidth}{!}{%
\begin{tabular}{llcccccc}
\toprule
Transform & Adaptation & DINO $\downarrow$ & Edge $\uparrow$ & \CLIPT $\uparrow$ & CSD $\uparrow$ & $\Delta E_{00}\downarrow$ & LPIPS $\downarrow$ \\
\midrule
NSST & Scale-only & 0.6413 & 0.902645 & \textbf{0.342469} & \textbf{0.558335} & 12.360673 & 0.434451 \\
NSST & Scale+Direction & \textbf{0.6274} & \textbf{0.904151} & 0.341949 & 0.557640 & \textbf{12.097185} & \textbf{0.425276} \\
DTCWT & Scale-only & 0.6674 & 0.899972 & 0.343685 & 0.539182 & 13.564509 & 0.454775 \\
DTCWT & Scale+Direction & \textbf{0.6536} & \textbf{0.903597} & \textbf{0.344153} & \textbf{0.542858} & \textbf{13.387058} & \textbf{0.449578} \\
DWT & Scale-only & 0.6517 & 0.902969 & \textbf{0.342219} & 0.541375 & 13.661466 & 0.450149 \\
DWT & Scale+Direction & \textbf{0.6495} & \textbf{0.903659} & 0.342041 & \textbf{0.545559} & \textbf{13.518173} & \textbf{0.447336} \\
NSCT & Scale-only & 0.6559 & \textbf{0.898018} & \textbf{0.343129} & 0.555116 & 11.812372 & 0.440765 \\
NSCT & Scale+Direction & \textbf{0.6554} & 0.897570 & 0.343015 & \textbf{0.555544} & \textbf{11.810691} & \textbf{0.440481} \\
\bottomrule
\end{tabular}}
\end{table}

\begin{table}[H]
\caption{SD2.1 transform ablation for appearance changing on Color40 ($n=40$) under the directional isolation setting described in Appendix~\ref{app:backbones}.}
\label{tab:sd21_change_transform}
\centering
\scriptsize
\begin{tabular}{llccc}
\toprule
Transform & Adaptation & DINO $\downarrow$ & Edge $\uparrow$ & \CLIPT $\uparrow$ \\
\midrule
NSST & Scale-only & 0.5905 & 0.901333 & \textbf{0.329739} \\
NSST & Scale+Direction & \textbf{0.5810} & \textbf{0.903617} & 0.329178 \\
DTCWT & Scale-only & 0.6038 & 0.897254 & \textbf{0.332482} \\
DTCWT & Scale+Direction & \textbf{0.6021} & \textbf{0.901915} & 0.332171 \\
DWT & Scale-only & 0.6001 & 0.897642 & 0.332141 \\
DWT & Scale+Direction & \textbf{0.5977} & \textbf{0.900713} & \textbf{0.332188} \\
NSCT & Scale-only & 0.5971 & 0.897179 & 0.327959 \\
NSCT & Scale+Direction & \textbf{0.5947} & \textbf{0.898957} & \textbf{0.328705} \\
\bottomrule
\end{tabular}
\end{table}

\FloatBarrier
\section{Targeted Assignment and Appearance-Alignment Controls}
\label{app:targeted_ablations}

The two controls in this section isolate specific mechanisms that are not answered by the broader architecture and transform sweeps. The first asks whether the directional controller benefits from assigning source-derived weights to the \emph{correct} bands, rather than from non-uniform weighting alone. The second isolates the task-specific contribution of \dcrci after AdaIN in the final appearance-preservation configuration.

\paragraph{Correct directional assignment.}
Table~\ref{tab:shuffled_direction} compares Scale-only, the proposed Scale+Direction assignment, and a Scale+Shuffled-Direction control on PIE-Bench Color40. All three use the same final SPECTRA-SD2.1 appearance-changing configuration: three-scale NSST ($J=3$), \ssmo, LF scheduling, and \selfattnconv are enabled, while AdaIN and \dcrci are disabled. Scale+Shuffled-Direction applies a fixed non-identity cyclic permutation to the direction-specific fusion weights within each scale, preserving the weight multiset while breaking its correspondence to the source directional bands. Correct assignment gives the best mean DINO and Edge preservation. The shuffled control yields slightly higher \CLIPT. Table~\ref{tab:direction_uncertainty} reports the paired uncertainty against both shuffled assignment and Scale-only.

\begin{table}[H]
\caption{Correct directional assignment on PIE-Bench Color40 ($n=40$). Scale+Shuffled-Direction preserves the direction-weight multiset within each scale but cyclically reassigns the weights to different bands. All other settings are identical. Bold denotes the best value per metric.}
\label{tab:shuffled_direction}
\centering
\scriptsize
\begin{tabular}{lccc}
\toprule
Adaptation & DINO $\downarrow$ & Edge $\uparrow$ & \CLIPT $\uparrow$ \\
\midrule
Scale-only & 0.1485 & 0.970097 & 0.310990 \\
Scale+Direction (correct) & \textbf{0.1483} & \textbf{0.970599} & 0.310811 \\
Scale+Shuffled-Direction & 0.1496 & 0.969944 & \textbf{0.311387} \\
\bottomrule
\end{tabular}
\end{table}

\begin{table}[H]
\addtocounter{table}{-1}
\renewcommand{\thetable}{H.1}
\renewcommand{\theHtable}{direction-paired}
\caption{Paired mean differences and 95\% confidence intervals for directional assignment on Color40. Each cell reports Correct minus the indicated control. Intervals use 10,000 paired bootstrap resamples of the 40 SD2.1 cases generated at the original seed.}
\label{tab:direction_uncertainty}
\centering
\footnotesize
\begin{tabular}{@{}lccc@{}}
\toprule
Comparison & $\Delta$ DINO $\downarrow$ & $\Delta$ Edge $\uparrow$ & $\Delta$ \CLIPT $\uparrow$ \\
\midrule
Correct $-$ Shuffled & \shortstack{$-0.001324$ \\ $[-0.002495, -0.000193]$} & \shortstack{$+0.000656$ \\ $[+0.000042, +0.001316]$} & \shortstack{$-0.000576$ \\ $[-0.000965, -0.000232]$} \\[0.4em]
Correct $-$ Scale-only & \shortstack{$-0.000211$ \\ $[-0.001868, +0.001629]$} & \shortstack{$+0.000503$ \\ $[-0.000032, +0.001091]$} & \shortstack{$-0.000178$ \\ $[-0.000546, +0.000200]$} \\
\bottomrule
\end{tabular}
\end{table}

\paragraph{Effect of DCRCI after AdaIN.}
Table~\ref{tab:dcrci_ablation} isolates \dcrci on PIE-Bench Style-9. Both variants use the final SPECTRA-SD2.1 preservation configuration with three-scale NSST ($J=3$), Scale+Direction adaptation, \ssmo, LF scheduling, \selfattnconv, and AdaIN; source--target pairs, cached DDIM trajectories, and seed are identical. The only difference is whether \dcrci is applied after AdaIN. Adding \dcrci improves CSD, reduces $\Delta E_{00}$ from 5.016290 to 4.306591, and slightly improves Edge similarity, consistent with its intended role as an internal source-relative chroma-correction stage. AdaIN alone retains better DINO, \CLIPT, and LPIPS, showing that \dcrci shifts the operating point toward source color/style fidelity rather than uniformly improving every objective.

\begin{table}[H]
\caption{Appearance-alignment ablation on PIE-Bench Style-9 ($n=80$). Both variants use identical SPECTRA-SD2.1 settings and differ only in whether \dcrci is applied after AdaIN. Bold denotes the better value per metric.}
\label{tab:dcrci_ablation}
\centering
\scriptsize
\resizebox{\linewidth}{!}{%
\begin{tabular}{lcccccc}
\toprule
Appearance alignment & DINO $\downarrow$ & Edge $\uparrow$ & \CLIPT $\uparrow$ & CSD $\uparrow$ & $\Delta E_{00}\downarrow$ & LPIPS $\downarrow$ \\
\midrule
AdaIN only & \textbf{0.1509} & 0.970445 & \textbf{0.313372} & 0.813432 & 5.016290 & \textbf{0.140420} \\
AdaIN + \dcrci & 0.1535 & \textbf{0.971377} & 0.303636 & \textbf{0.830143} & \textbf{4.306591} & 0.161643 \\
\bottomrule
\end{tabular}}
\end{table}

\FloatBarrier
\section{Additional Directional-Heterogeneity Evidence}
\label{app:directional}

\begin{figure}[!t]
\centering
\begin{subfigure}[t]{0.49\linewidth}
    \centering
    \includegraphics[width=\linewidth]{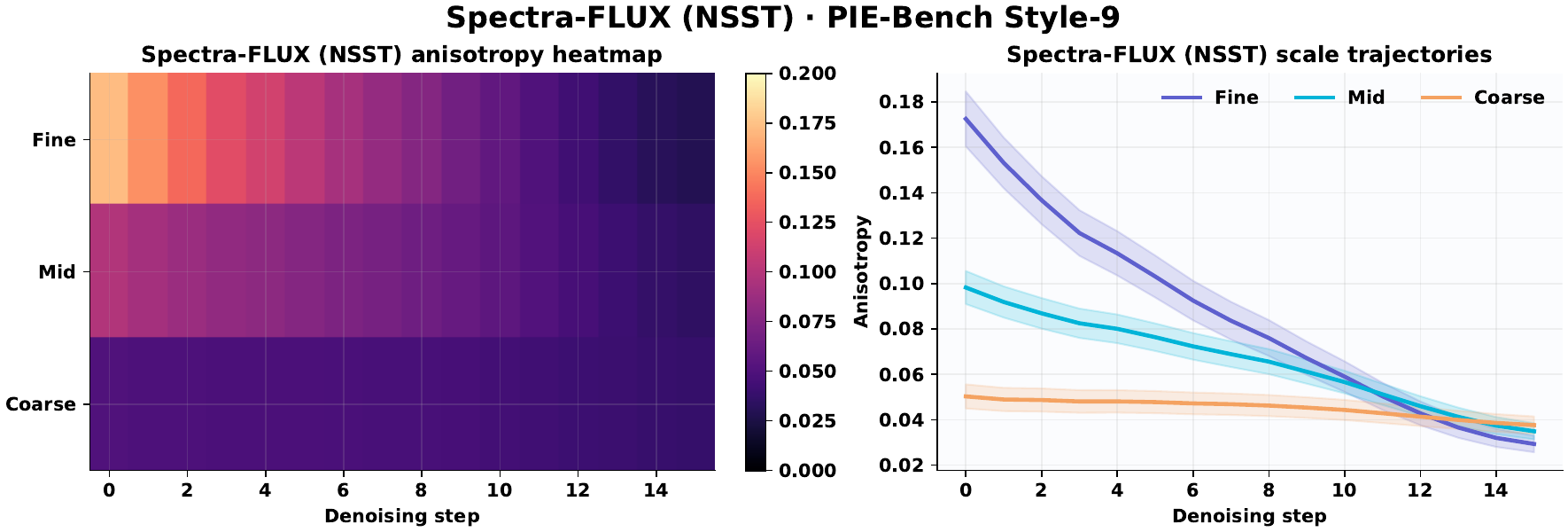}
    \caption{Anisotropy over the native FlowMatch trajectory.}
\end{subfigure}\hfill
\begin{subfigure}[t]{0.49\linewidth}
    \centering
    \includegraphics[width=\linewidth]{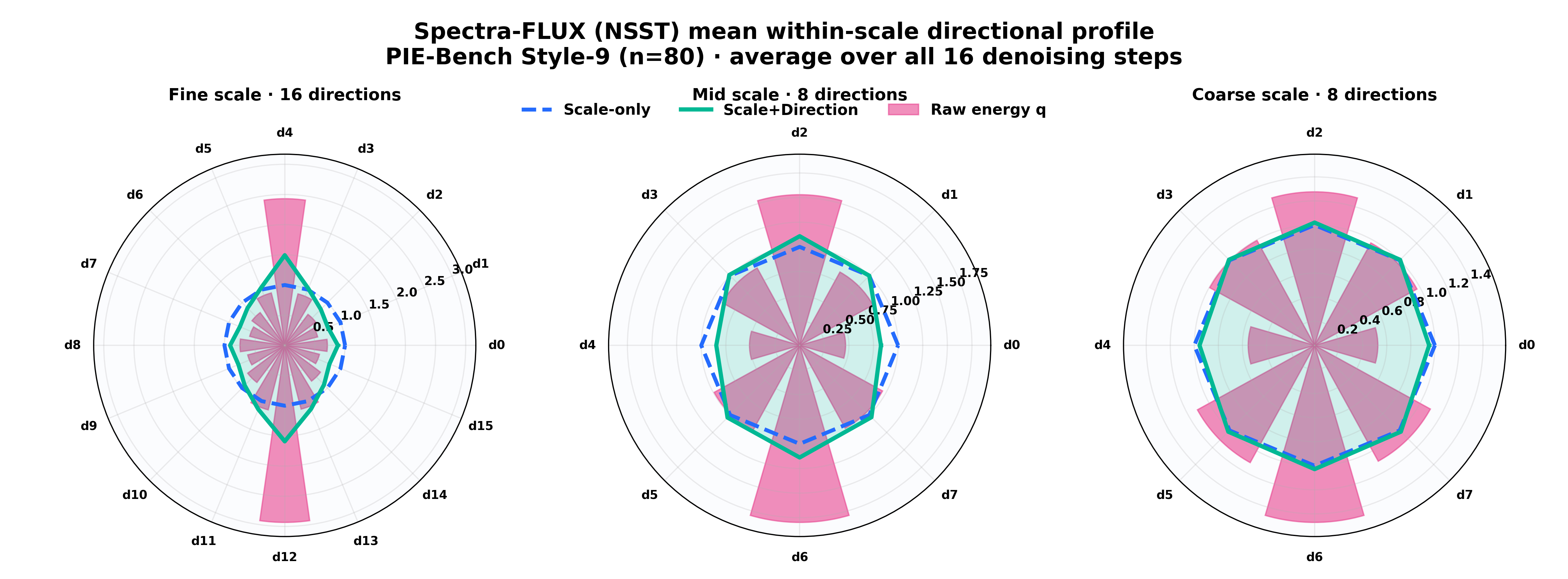}
    \caption{Mean within-scale directional treatment.}
\end{subfigure}
\vspace{-1mm}
\begin{subfigure}[t]{0.78\linewidth}
    \centering
    \includegraphics[width=\linewidth]{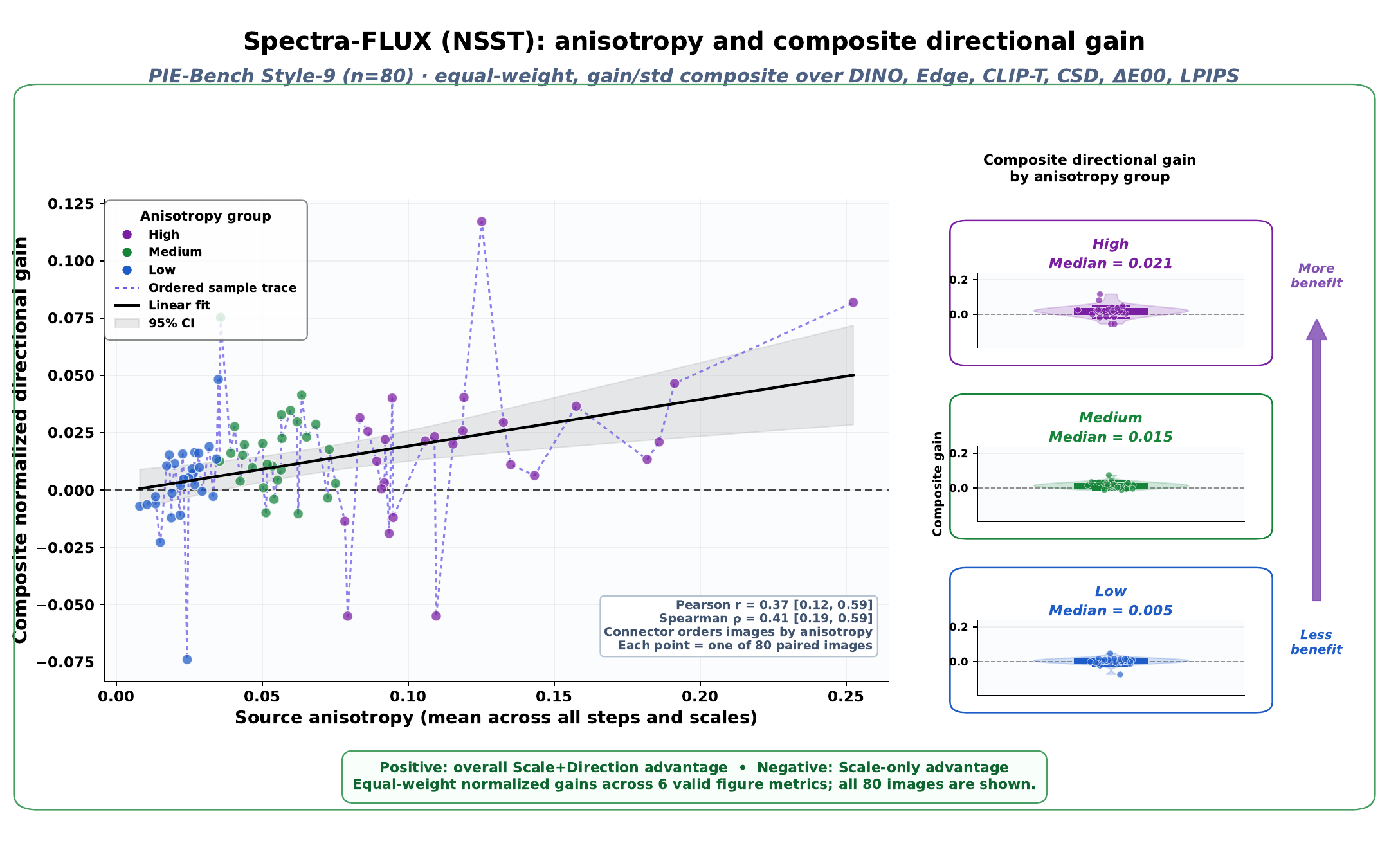}
    \caption{Source anisotropy versus paired directional gain.}
\end{subfigure}
\vspace{-1mm}
\caption{Mechanistic analysis for \method (FLUX.1-dev, NSST) on Style-9. (a) Source anisotropy remains scale-dependent and time-varying under the selected native FlowMatch trajectory. (b) Scale+Direction produces non-uniform source-dependent allocation while Scale-only is direction-invariant. (c) The anisotropy--gain association uses the same signed, standardized composite as Fig.~\ref{fig:mechanism_main}(c); it is descriptive rather than causal.}
\label{fig:flux_mechanism}
\end{figure}

\begin{figure}[!t]
\centering
\begin{subfigure}[t]{0.49\linewidth}
    \centering
    \includegraphics[width=\linewidth]{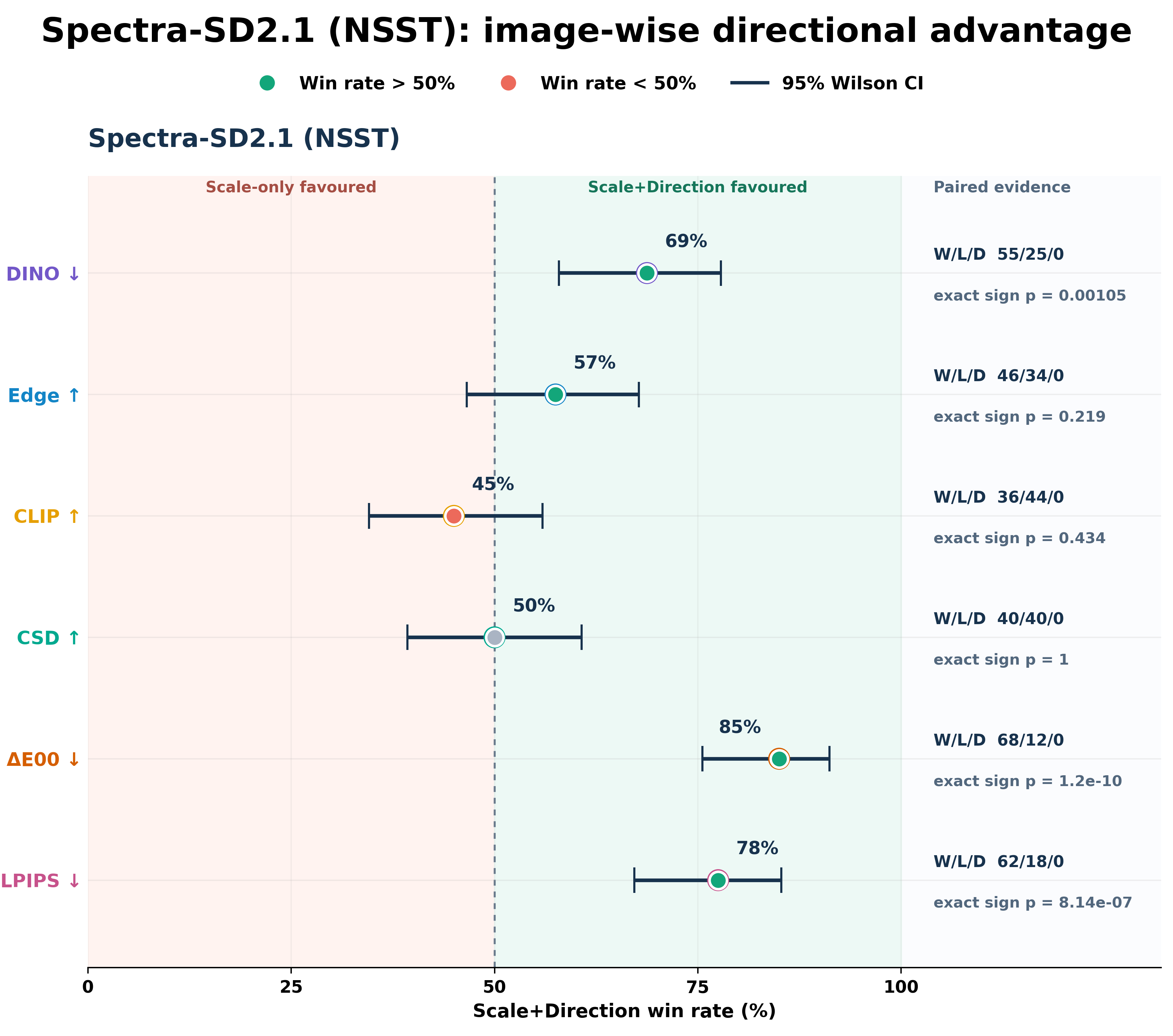}
    \caption{SD2.1.}
\end{subfigure}\hfill
\begin{subfigure}[t]{0.49\linewidth}
    \centering
    \includegraphics[width=\linewidth]{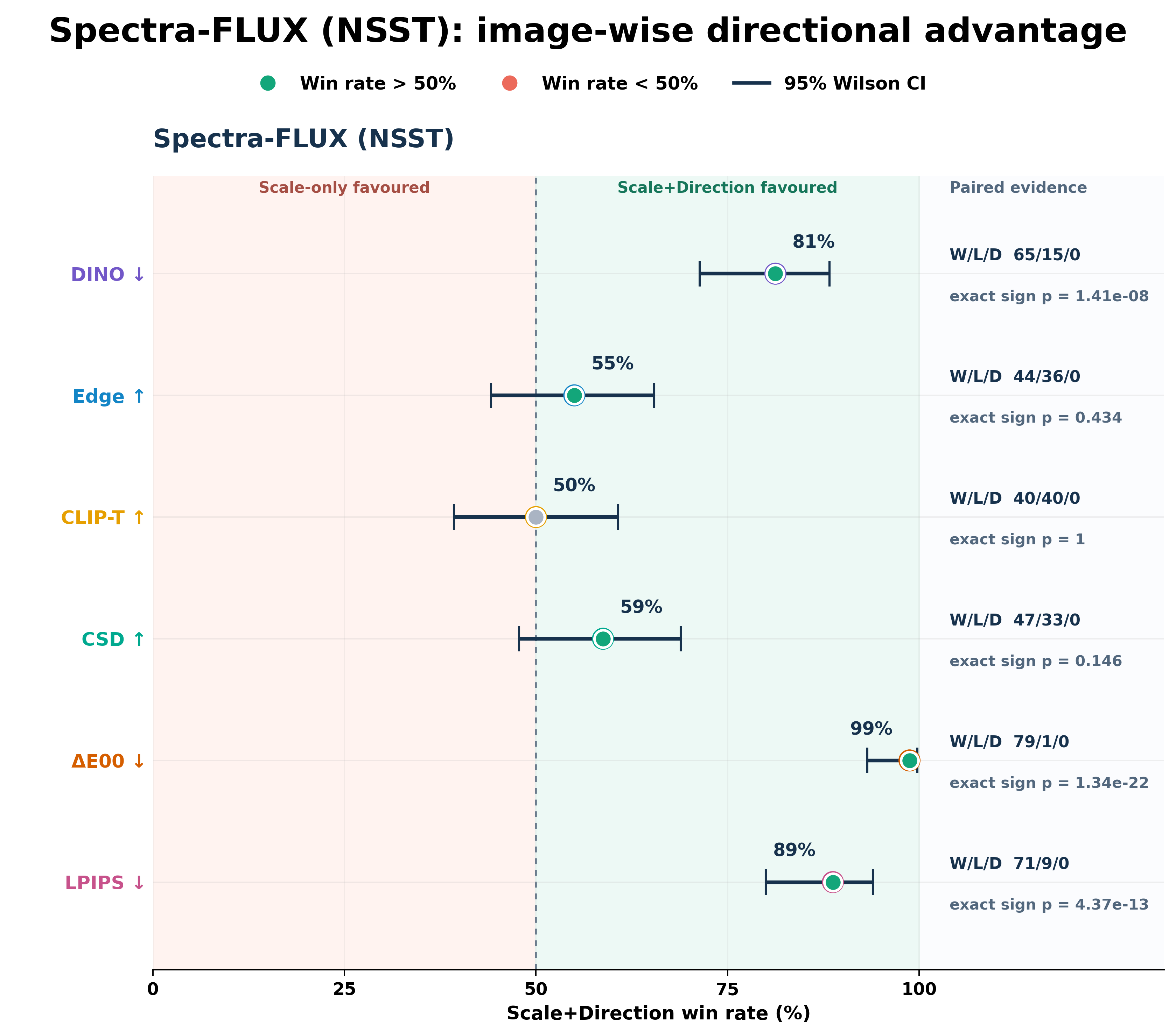}
    \caption{FLUX.1-dev.}
\end{subfigure}
\vspace{-1mm}
\caption{Image-wise directional advantage over Scale-only on Style-9. Each point is the Scale+Direction win fraction for one metric; horizontal bars are 95\% Wilson intervals, and exact two-sided sign tests compare wins with losses while excluding draws. The case-wise view shows that directional refinement is strong for several preservation metrics but mixed or neutral for others, matching the qualified claim in the main text.}
\label{fig:casewise_wins}
\end{figure}

The main paper reports the SD2.1 mechanistic sequence. We repeat the same measurements for FLUX.1-dev and keep the two backbones separate. Figure~\ref{fig:flux_mechanism} mirrors the three main-paper diagnostics: the source signal changes across scale and time, the controller responds to the native directional-energy profile, and higher anisotropy is associated with larger paired directional gain. Figure~\ref{fig:casewise_wins} complements average metrics with case-wise win rates and uncertainty.

\FloatBarrier
\section{Additional Qualitative Results}
\label{app:qualitative}

\begin{figure}[!t]
\centering
\includegraphics[width=\linewidth]{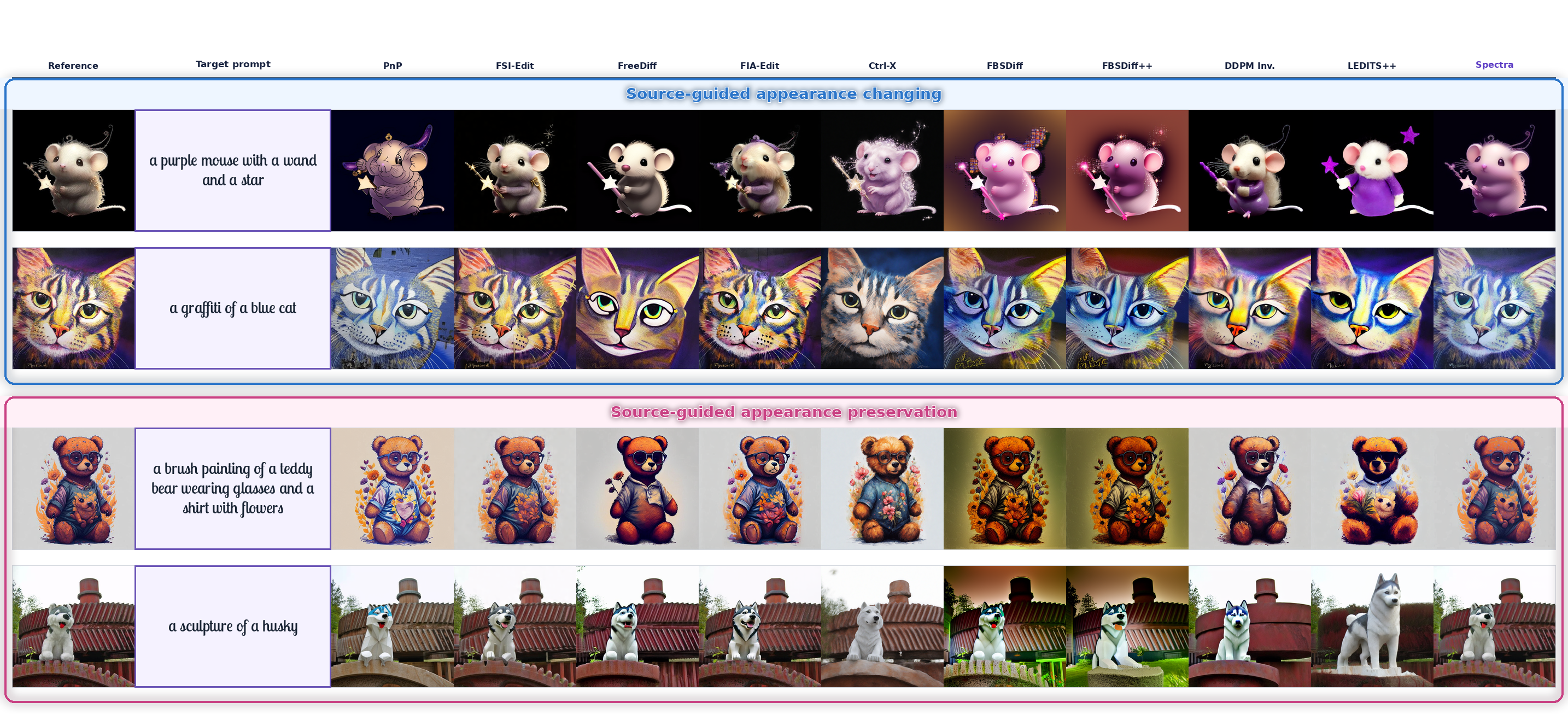}
\caption{Additional baseline comparisons for appearance-changing and appearance-preserving edits. These examples complement Fig.~\ref{fig:qualitative} with different sources and target prompts.}
\label{fig:appendix_baselines}
\end{figure}

\begin{figure}[!t]
\centering
\includegraphics[width=\linewidth]{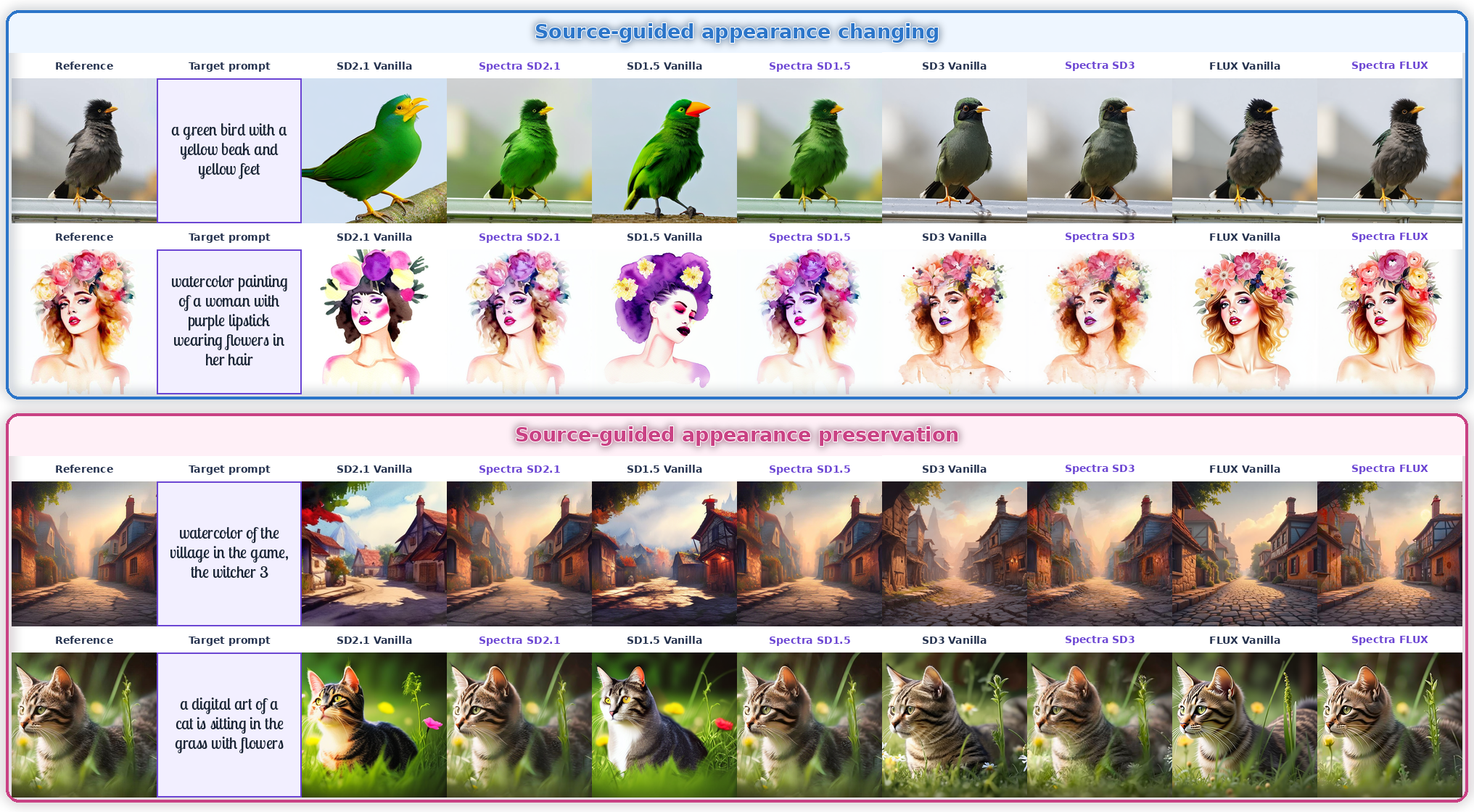}
\caption{Additional Vanilla-versus-\method comparisons across diffusion and flow backbones. The figure is intended as a cross-backbone qualitative check rather than a controlled component ablation because backbone-native source construction and the selected operating points differ.}
\label{fig:appendix_backbones}
\end{figure}

\FloatBarrier
\section{Implementation and Reproducibility Details}
\label{app:repro}

All experiments were run on a Linux server with four NVIDIA GeForce RTX 3090 GPUs, each with 24 GB VRAM, using NVIDIA driver 580.173.02. A generation or evaluation worker uses one GPU. Multi-GPU execution only shards independent images or experiment conditions; there is no cross-device activation exchange or gradient aggregation. Each controlled model ablation fixes the source image, target prompt, seed, backbone checkpoint, inference schedule, and resolution except for the factors being tested. Generation and evaluation artifacts are stored per sample with explicit identifiers, and an experiment is considered complete only when every expected sample is present. Missing samples are not silently skipped.

The common evaluator pins the DINO checkpoint and repository state, the official CSD ViT-L checkpoint, CLIP/HED preprocessing, CIEDE2000 validation data, and official LPIPS-AlexNet v0.1. Each run logs transform metadata, scale and directional energies, attention statistics, SSMO gate statistics, reconstruction diagnostics, model revision, scheduler settings, and appearance-alignment diagnostics. These records are intended to make the scale/direction mechanism auditable independently of the rendered image.

\subsection{Computational cost}
\label{app:runtime}

We profiled batch-size-one editing on an RTX 3090. First-edit latency includes source preparation, editing, and final decoding and excludes model loading and image writing. For SD2.1, we also report inversion time and the latency of a subsequent edit that reuses the cached inversion. Peak GPU memory is reported in GiB. Dashes indicate an unreported timing breakdown.

\begin{table}[H]
\caption{Computational cost on PIE-Bench Color40 ($n=40$). Timings (s) and per-case peak VRAM (GiB) are averaged over cases. FIA-Edit uses its native SD3.5-Medium backbone at the same resolution as the SD3-Medium runs.}
\label{tab:cost_color}
\centering
\footnotesize
\setlength{\tabcolsep}{4pt}
\begin{tabular}{@{}lllrrrr@{}}
\toprule
Method & Backbone & Resolution & \shortstack{First\\edit (s)} & \shortstack{Inversion\\(s)} & \shortstack{Cached\\edit (s)} & \shortstack{Peak VRAM\\(GiB)} \\
\midrule
SPECTRA & SD2.1 & 512$\times$512 & 96.78 & 80.97 & 15.76 & 4.51 \\
Vanilla & SD2.1 & 512$\times$512 & 82.71 & 80.26 & 2.45 & 3.12 \\
PnP & SD2.1 & 512$\times$512 & 107.69 & 96.53 & 11.19 & 4.51 \\
SPECTRA & SD3-Medium & 1024$\times$1024 & 36.96 & --- & --- & 18.78 \\
Vanilla & SD3-Medium & 1024$\times$1024 & 11.35 & --- & --- & 18.76 \\
FIA-Edit & SD3.5-Medium & 1024$\times$1024 & 41.34 & --- & --- & 20.09 \\
\bottomrule
\end{tabular}
\end{table}

\begin{table}[H]
\caption{Computational cost on PIE-Bench Style-9 ($n=80$) using the same profiling protocol. Timings and per-case peak VRAM are averaged over cases. Full appearance alignment uses AdaIN and \dcrci. The AdaIN-only runs disable \dcrci and its iterative VAE cycle. FIA-Edit uses native SD3.5-Medium.}
\label{tab:cost_style}
\centering
\footnotesize
\setlength{\tabcolsep}{4pt}
\begin{tabular}{@{}llllrrrr@{}}
\toprule
Method & Backbone & Appearance & Resolution & \shortstack{First\\edit (s)} & \shortstack{Inversion\\(s)} & \shortstack{Cached\\edit (s)} & \shortstack{Peak VRAM\\(GiB)} \\
\midrule
SPECTRA & SD2.1 & AdaIN + \dcrci & 512$\times$512 & 110.57 & 82.31 & 28.30 & 4.52 \\
SPECTRA & SD2.1 & AdaIN only & 512$\times$512 & 97.57 & 80.16 & 17.45 & 4.52 \\
Vanilla & SD2.1 & --- & 512$\times$512 & 83.69 & 81.26 & 2.43 & 3.12 \\
PnP & SD2.1 & --- & 512$\times$512 & 107.53 & 96.50 & 11.04 & 4.51 \\
SPECTRA & SD3-Medium & AdaIN + \dcrci & 1024$\times$1024 & 55.73 & --- & --- & 19.49 \\
SPECTRA & SD3-Medium & AdaIN only & 1024$\times$1024 & 37.78 & --- & --- & 18.78 \\
Vanilla & SD3-Medium & --- & 1024$\times$1024 & 11.86 & --- & --- & 18.76 \\
FIA-Edit & SD3.5-Medium & Native & 1024$\times$1024 & 42.36 & --- & --- & 20.09 \\
\bottomrule
\end{tabular}
\end{table}

On Color40, \method completes the first edit 10.91~s faster than PnP on SD2.1 at the same peak VRAM, primarily due to shorter inversion. Its cached edit takes 15.76~s, compared with 11.19~s for PnP and 2.45~s for Vanilla. At $1024\times1024$, SD3-Medium \method completes the first edit 4.38~s faster than native SD3.5-Medium FIA-Edit and uses 1.31~GiB less peak VRAM. Compared with Vanilla on the same backbone, first-edit and peak-memory overhead on Color40 are 14.07~s and 1.39~GiB on SD2.1, and 25.61~s and 0.02~GiB on SD3-Medium. On Style-9, the corresponding overheads with full \method are 26.88~s and 1.40~GiB on SD2.1, and 43.87~s and 0.73~GiB on SD3-Medium. Disabling \dcrci reduces first-edit latency by 13.00~s on SD2.1 and 17.95~s on SD3-Medium. With AdaIN only, \method completes the first edit faster than SD2.1 PnP and than resolution-matched FIA-Edit. Section~\ref{sec:tradeoff} gives within-backbone overhead over Vanilla.

\FloatBarrier

\section{Controlled Validation on SD2.1}
\label{app:controlled_sd}

\subsection{Full-controller factorial ablation}
\label{app:factorial}

We independently toggle joint source self-attention and convolutional feature injection ($I$) and the complete spectral controller ($C$). Setting either factor to zero disables the corresponding intervention. When enabled, injection uses $(a_t,f_t)=(0.8,1.0)$. The controller combines LF fusion, adaptive scale--direction weighting, \ssmo, and spectral reconstruction. Source construction and sampling settings are shared by all four conditions. AdaIN and \dcrci are off throughout Color40 and on throughout Style-9.

\begin{table}[H]
\caption{Matched factorial ablation of source feature injection and the spectral controller on SD2.1. Values are means over the same image--prompt pairs in each task. Bold marks the best mean per metric.}
\label{tab:controller_factorial}
\centering
\footnotesize
\textbf{(a) Color40 ($n=40$)}\par\smallskip
\begin{tabular}{@{}lrrr@{}}
\toprule
Condition & DINO $\downarrow$ & Edge $\uparrow$ & \CLIPT $\uparrow$ \\
\midrule
Neither ($I=0$, $C=0$) & 0.73496 & 0.88620 & \textbf{0.33317} \\
Injection only ($I=1$, $C=0$) & 0.19938 & 0.96319 & 0.31689 \\
Controller only ($I=0$, $C=1$) & 0.58096 & 0.90362 & 0.32918 \\
Full \method ($I=1$, $C=1$) & \textbf{0.14832} & \textbf{0.97060} & 0.31081 \\
\bottomrule
\end{tabular}

\medskip
\textbf{(b) Style-9 ($n=80$)}\par\smallskip
\begin{tabular}{@{}lrrrrrr@{}}
\toprule
Condition & DINO $\downarrow$ & Edge $\uparrow$ & \CLIPT $\uparrow$ & CSD $\uparrow$ & $\Delta E_{00}\downarrow$ & LPIPS $\downarrow$ \\
\midrule
Neither ($I=0$, $C=0$) & 0.55899 & 0.91749 & \textbf{0.33194} & 0.60665 & 8.95588 & 0.39912 \\
Injection only ($I=1$, $C=0$) & 0.22486 & 0.96430 & 0.31175 & 0.76792 & 5.16615 & 0.21300 \\
Controller only ($I=0$, $C=1$) & 0.41180 & 0.93488 & 0.32177 & 0.69717 & 6.51298 & 0.30725 \\
Full \method ($I=1$, $C=1$) & \textbf{0.15350} & \textbf{0.97138} & 0.30364 & \textbf{0.83014} & \textbf{4.30659} & \textbf{0.16164} \\
\bottomrule
\end{tabular}
\end{table}

The controller improves DINO and Edge at both injection levels. With injection off, adding it lowers mean DINO by 0.15400 and 0.14719 and raises Edge by 0.01742 and 0.01739 on Color40 and Style-9, respectively. Section~\ref{sec:controlled_validation} reports the incremental effect when injection is held on. These comparisons measure the effect of the full controller at each injection setting.

\FloatBarrier
\subsection{SSMO spatial allocation and temporal weighting}
\label{app:gate_controls}

Starting from full SD2.1 \method, we record the reference gate $G^{\mathrm{ref}}_{k,t}(p)$ and scale--direction weight $w^{\mathrm{ref}}_{k,t}$ for each case. Here $k$ indexes a transform atom, $t$ a denoising step, and $p$ a spatial position. We compare the original setting with three controls:
\begin{itemize}
\item \textbf{Mean-matched constant gate.} We use $G^{\mathrm{const}}_{k,t}(p)=\operatorname{mean}_{p'}G^{\mathrm{ref}}_{k,t}(p')$, preserving the spatial mean separately for each atom and step.
\item \textbf{Spatially shuffled gate.} We use $G^{\mathrm{shuffle}}_{k,t}(p)=G^{\mathrm{ref}}_{k,t}(\pi(p))$, where $\pi$ permutes spatial positions. This changes where the gate values are applied while preserving their distribution.
\item \textbf{Time-averaged weights.} We use $\bar w_k=T^{-1}\sum_{t=1}^{T}w^{\mathrm{ref}}_{k,t}$ at each step before \ssmo, where $T$ is the number of denoising steps. Target attention is recomputed at each step.
\end{itemize}

Source construction, feature injection, sampling settings, and task-specific appearance correction are fixed across conditions. The terminal low-pass source gate stays zero.

\begin{table}[H]
\caption{Spatial gate and temporal weight controls on SD2.1. AdaIN and \dcrci are disabled on Color40 and enabled on Style-9. Values are dataset means. Bold marks the best mean per metric and task.}
\label{tab:spatial_temporal_controls}
\centering
\footnotesize
\textbf{(a) Color40 ($n=40$)}\par\smallskip
\begin{tabular}{@{}lrrr@{}}
\toprule
Condition & DINO $\downarrow$ & Edge $\uparrow$ & \CLIPT $\uparrow$ \\
\midrule
Original \method & \textbf{0.14832} & \textbf{0.97060} & 0.31081 \\
Mean-matched constant gate & 0.14853 & 0.97018 & 0.31078 \\
Spatially shuffled gate & 0.14841 & 0.97007 & 0.31071 \\
Time-averaged weights & 0.14964 & 0.97045 & \textbf{0.31103} \\
\bottomrule
\end{tabular}

\medskip
\textbf{(b) Style-9 ($n=80$)}\par\smallskip
\begin{tabular}{@{}lrrrrrr@{}}
\toprule
Condition & DINO $\downarrow$ & Edge $\uparrow$ & \CLIPT $\uparrow$ & CSD $\uparrow$ & $\Delta E_{00}\downarrow$ & LPIPS $\downarrow$ \\
\midrule
Original \method & \textbf{0.15350} & \textbf{0.97138} & 0.30364 & 0.83014 & \textbf{4.30659} & \textbf{0.16164} \\
Mean-matched constant gate & 0.15384 & 0.97066 & 0.30369 & 0.82834 & 4.32074 & 0.16253 \\
Spatially shuffled gate & 0.15411 & 0.97047 & 0.30373 & 0.82871 & 4.32472 & 0.16263 \\
Time-averaged weights & 0.15499 & 0.97136 & \textbf{0.30374} & \textbf{0.83031} & 4.30960 & 0.16275 \\
\bottomrule
\end{tabular}
\end{table}

The original gate's mean DINO and Edge gains over both spatial controls are small and consistent across the two datasets. These contrasts support a role for the spatial distribution of the gate at fixed mean gate strength. Original \method also achieves the lowest $\Delta E_{00}$ and LPIPS on Style-9. With live target attention in both conditions, per-step weights lower mean DINO by 0.00132 on Color40 and 0.00149 on Style-9 compared with time-averaged weights.

\FloatBarrier
\section{Blinded Human Evaluation on Style-9}
\label{app:human_eval}

We compared balanced \method (SD2.1, one NSST scale, AdaIN enabled, and \dcrci disabled) with FIA-Edit. Their similar mean DINO across all 80 Style-9 image--prompt pairs motivated the comparison (Table~\ref{tab:tradeoff}). Eighteen participants evaluated 20 uniformly sampled Style-9 cases. Each trial displayed the source, the exact target instruction, and two anonymous outputs in randomized left--right order. Participants could select either output, both, or neither. An output counted as a success for its method if the participant judged that it clearly applied the requested rendering style while preserving the source shapes, poses, and layout. We computed both success proportions over the 360 paired judgments. To estimate 95\% confidence intervals for the two rates and their difference, we used 10,000 two-way bootstrap resamples over cases and participants, keeping the methods paired.

\begin{table}[H]
\caption{Blinded evaluation of joint editing and structural preservation on Style-9. Each method's rate is the proportion of paired judgments in which its output met both criteria. The paired difference is in percentage points. Confidence intervals resample cases and participants.}
\label{tab:human_eval}
\centering
\small
\begin{tabular}{@{}lcc@{}}
\toprule
Method or contrast & Joint success & 95\% CI \\
\midrule
FIA-Edit & 112/360 (31.1\%) & $[17.8\%,44.7\%]$ \\
Balanced \method & 135/360 (37.5\%) & $[23.9\%,51.7\%]$ \\
\method{} $-$ FIA-Edit & $+6.4$ pp & $[-14.7,+27.5]$ pp \\
\bottomrule
\end{tabular}
\end{table}

Balanced \method has the higher observed joint success rate, and the paired interval spans zero. The endpoint combines visible edit success and source geometry in one judgment. Preservation equivalence on the 20 surveyed cases remains open.

\end{document}

%% file: math_commands.tex
\usepackage{amsmath,amsfonts,bm}

\def\eqref#1{equation~\ref{#1}}

\def\1{\bm{1}}

\DeclareMathAlphabet{\mathsfit}{\encodingdefault}{\sfdefault}{m}{sl}
\SetMathAlphabet{\mathsfit}{bold}{\encodingdefault}{\sfdefault}{bx}{n}

